\documentclass[12pt,letterpaper]{article}

\usepackage[margin=1in]{geometry}
\usepackage{multirow}
\usepackage{amsmath,amssymb,amsfonts}
\usepackage{xcolor}
\usepackage{textcomp}
\usepackage{booktabs}
\usepackage{url}
\usepackage{hyperref}
\usepackage{graphicx}
\usepackage[numbers,sort&compress]{natbib}

\newcommand{\aimonly}[1]{}

\title{Bit-Identical Medical Deep Learning via Structured Orthogonal Initialization}

\author{Yakov P. Shkolnikov\\
Independent Researcher, Mountain View, CA, United States\\
\texttt{yshkolni@gmail.com}}

\date{}

\begin{document}

\maketitle

\begin{abstract}
Deep learning training is non-deterministic: identical code with different random seeds produces models that agree on aggregate metrics but disagree on individual predictions, with per-class AUC swings exceeding 20 percentage points on rare clinical classes. We present a framework for verified bit-identical training that eliminates three sources of randomness: weight initialization (via structured orthogonal basis functions), batch ordering (via golden ratio scheduling), and non-deterministic GPU operations (via architecture selection and custom autograd). The pipeline produces MD5-verified identical trained weights across independent runs.

On PTB-XL ECG rhythm classification, structured initialization significantly exceeds Kaiming across two architectures ($n\!=\!20$; Conformer $p = 0.016$, Baseline $p < 0.001$), reducing aggregate variance by $2$--$3{\times}$ and reducing per-class variability on rare rhythms by up to $7.5{\times}$ (TRIGU range: 4.1pp vs 30.9pp under Kaiming, independently confirmed by 3-fold CV). A four-basis comparison at $n{=}20$ shows all structured orthogonal bases produce equivalent performance (Friedman $p{=}0.48$), establishing that the contribution is deterministic structured initialization itself, not any particular basis function. Cross-domain validation on seven MedMNIST benchmarks ($n\!=\!20$, all $p > 0.14$) confirms no performance penalty on standard tasks; per-class analysis on imbalanced tasks (ChestMNIST, RetinaMNIST) shows the same variance reduction on rare classes observed in ECG. Cross-dataset evaluation on three external ECG databases confirms zero-shot generalization ($>$0.93 AFIB AUC).

\end{abstract}



\section{Introduction}\label{sec:introduction}

Deep learning for medical classification, including electrocardiogram (ECG) rhythm detection, histopathology tissue typing, and dermatoscopic lesion classification, has achieved expert-level performance on benchmark datasets~\cite{wagner2020ptbxl,strodthoff2021ptbxl,hannun2019cardiologist,ribeiro2020automatic}. Yet a barrier to clinical deployment persists: irreproducibility at the individual patient level.

\begin{figure}[!t]
\centering
\includegraphics[width=\linewidth]{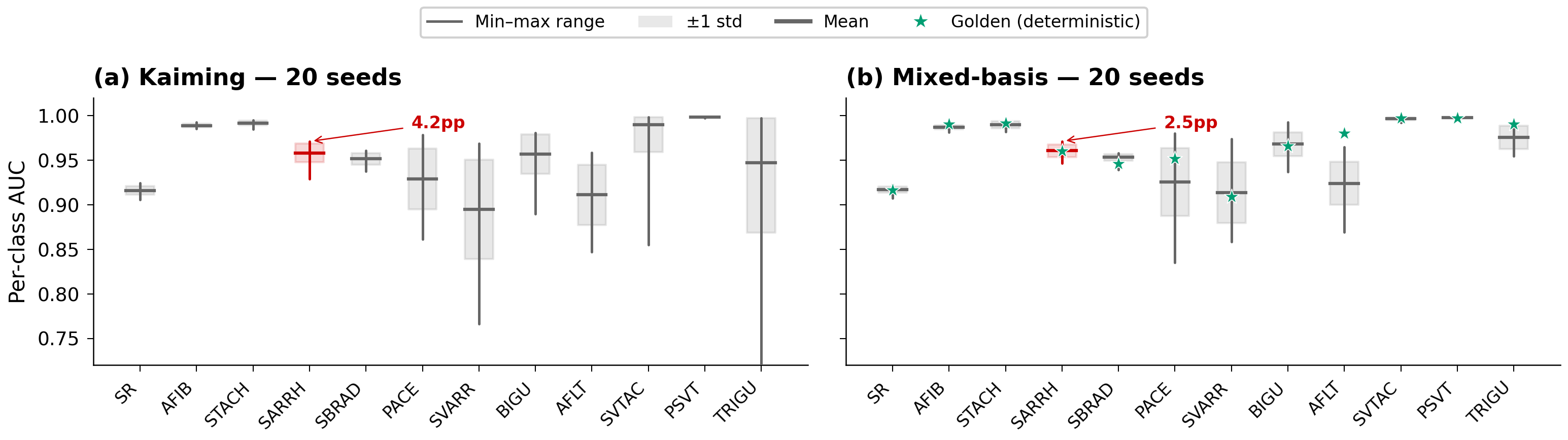}
\caption{Per-class performance variability (Conformer, $n\!=\!20$ seeds). (a)~Kaiming initialization: SARRH ($n_\text{pos}\!=\!77$) ranges 4.2pp across seeds despite being moderately represented. SVARR ($n_\text{pos}\!=\!14$) ranges 20.2pp; TRIGU ($n_\text{pos}\!=\!2$) ranges 30.9pp. (b)~Mixed-basis initialization: systematically tighter ranges. SARRH reduced from 4.2 to 2.5pp, SVARR from 20.2 to 11.5pp, TRIGU from 30.9 to 4.1pp ($7.5{\times}$). Green stars show the fully deterministic golden ratio run (zero variance by construction); golden falls within or above the seeded range for all classes. Thick horizontal lines: mean; shaded regions: $\pm$1 std; thin whiskers: full min--max range. Test-set positive counts (fold~10): SR\,1674, AFIB\,152, STACH\,82, SARRH\,77, SBRAD\,64, PACE\,28, SVARR\,14, BIGU\,8, AFLT\,7, SVTAC\,3, PSVT\,2, TRIGU\,2.}\label{fig:perclass}
\end{figure}

Standard deep learning training involves three sources of randomness, namely weight initialization (Kaiming uniform sampling~\cite{he2015delving}), stochastic batch ordering, and non-deterministic GPU operations (e.g., atomicAdd in pooling backward passes). Together, these mean that running identical code with different random seeds can produce models that agree on aggregate metrics but disagree on which patients they misdiagnose. In our ECG experiments, Kaiming-initialized Conformers across 20 initialization seeds exhibit per-class AUC swings of 20.2 percentage points for supraventricular arrhythmia (SVARR: 0.766--0.969, $n_\text{pos}\!=\!14$ test cases) and 11.8 points for paced rhythm (PACE: 0.861--0.979, $n_\text{pos}\!=\!28$), while aggregate macro AUC varies by only 4.1 percentage points (Fig.~\ref{fig:perclass}). A clinician reviewing two models trained from the same code with different seeds would encounter different failure modes: one model might miss SVARR, another might miss PACE, with no way to predict which patients are at risk from which model. Per-class instability of this magnitude is invisible in aggregate metrics.

Reproducibility is increasingly recognized as important in medical AI development. The FDA's AI/ML Software as a Medical Device Action Plan~\cite{fda2021aiml,fda2025lifecycle} and the EU AI Act (Regulation 2024/1689)~\cite{euaiact2024} discuss reproducibility and traceability in their guidance on good machine learning practices. Deterministic training directly supports these goals by enabling verifiable, auditable training pipelines, though the regulatory implications of algorithmic determinism have not been formally evaluated. Prior work on deterministic training has focused on GPU operation determinism (PyTorch's deterministic mode~\cite{paszke2019pytorch}) or seed fixing, but these approaches merely make randomness reproducible given a specific seed, and they do not eliminate the seed dependence itself. Recent work has established that structured initialization can match random Kaiming initialization~\cite{zhao2022zero,miravet2025sinusoidal} (Section~\ref{sec:related}), confirming that random weights are unnecessary. These methods do not address the remaining sources of non-determinism (batch ordering and GPU operations) that prevent fully reproducible training. Our contribution is a complete determinism framework that eliminates all three sources simultaneously, not any particular basis function.

We achieve verified bit-identical deep learning training by systematically eliminating all three sources of randomness. Weight initialization uses structured orthogonal basis functions (DCT, Hadamard, Hartley) computed analytically, producing identical weights on every run without any random seed. Batch ordering uses golden ratio scheduling, a deterministic method requiring no random seed. Non-deterministic GPU operations are eliminated by architectural choice (the Conformer uses deterministic implementations for all operations) or custom autograd functions. Together, these components yield trained models with bit-identical weights across independent runs on the same hardware, verified by MD5 hash comparison.

Three levels of determinism are distinguished. \emph{Initialization determinism} produces identical weights regardless of seed, and all architectures in this study attain it. \emph{Full training determinism} additionally requires deterministic batch ordering and deterministic GPU operations. Only the Conformer with golden ratio batch ordering attains full training determinism, producing MD5-verified bit-identical trained weights. \emph{Inference determinism}, where the same trained model and input produce the same output, holds trivially for all models.
This paper makes five contributions:
\begin{itemize}
\item MD5-verified bit-identical training on medical classification tasks in both 1D (ECG) and 2D (medical imaging).
\item Variance decomposition separating initialization variance (eliminated by construction), batch-ordering variance (std\,=\,0.005), and fold/data variance (std\,=\,0.012).
\item A four-basis comparison (DCT, Hadamard, Hartley, sinusoidal, $n{=}20$ each) demonstrating equivalent mean performance (Friedman $p{=}0.48$), establishing that the contribution is deterministic structured initialization itself, not any particular basis function.
\item A $2{\times}2$ golden-seed design illustrating seed-dependent per-class variation under Kaiming initialization (e.g., trigeminy AUC 0.781 vs 0.980 with structured init).
\item Validation across two ECG architectures, seven MedMNIST medical image benchmarks ($n{=}20$ per condition), CIFAR-100, and three external ECG databases.
\end{itemize}

\section{Related Work}\label{sec:related}

\subsection{Weight initialization}
Glorot and Bengio~\cite{glorot2010understanding} derived variance-preserving conditions for sigmoid/tanh activations. He et al.~\cite{he2015delving} extended this to ReLU networks with Kaiming initialization, now the default in most deep learning frameworks. Saxe et al.~\cite{saxe2014exact} showed that orthogonal weight matrices enable depth-independent learning dynamics and faithful gradient propagation. Pennington et al.~\cite{pennington2017resurrecting} showed that orthogonal initialization achieves \emph{dynamical isometry}, the concentration of all singular values of the input-output Jacobian near unity, enabling deep signal propagation through sigmoid networks. Xiao et al.~\cite{xiao2018dynamical} extended dynamical isometry to convolutional networks, enabling training of 10,000-layer vanilla CNNs with orthogonal initialization. Hu et al.~\cite{hu2020provable} provided the first provable convergence benefit for orthogonal initialization in deep linear networks, establishing that orthogonal weights accelerate optimization, not merely preserve signal norms. Mishkin and Matas~\cite{mishkin2016lsuv} proposed Layer-Sequential Unit-Variance (LSUV) initialization, demonstrating that careful initialization can replace batch normalization~\cite{ioffe2015batch}. Zhang et al.~\cite{zhang2019fixup} introduced Fixup initialization, enabling stable training of very deep ResNets through residual branch rescaling. All of these methods remain stochastic. Different random seeds produce different weight matrices, even when variance is properly controlled.

\subsection{Deterministic and structured initialization}
Several recent works have explored non-random initialization. Zhao et al.~\cite{zhao2022zero} proposed ZerO initialization using Hadamard matrices to initialize networks with only zeros and ones, demonstrating on ImageNet that random weights are unnecessary. Fernandez-Hernandez et al.~\cite{miravet2025sinusoidal} introduced sinusoidal initialization, matching Kaiming accuracy across CNNs and vision transformers with faster convergence. Aghajanyan~\cite{aghajanyan2017convolution} proposed convolution-aware initialization by constructing orthogonal filters in the Fourier domain and inverting to standard space, providing an early demonstration that exploiting convolution structure yields improvements over random initialization. Ulicny et al.~\cite{ulicny2022harmonic} developed Harmonic Convolutional Networks using DCT basis functions as preset spectral filters, demonstrating DCT energy compaction for network compression. Our work differs from Ulicny et al.\ in that they replace convolutional layers entirely with fixed DCT filter banks that learn linear combinations of basis functions, whereas we initialize standard convolutional layers with DCT-derived weights that are then freely updated by gradient descent. Our approach differs from ZerO initialization in using DCT bases (which provide smooth frequency-domain structure suited to quasi-periodic signals like ECGs) rather than binary Hadamard patterns, and in supporting multiple basis families (DCT, Hadamard, Hartley, sinusoidal) that can be used individually or combined across network stages. To our knowledge, Hartley basis functions have not previously been used for neural network weight initialization. Pan et al.~\cite{pan2025idinit} proposed IDInit using padded identity-like matrices, demonstrating at ICLR 2025 that structured initialization can overcome rank constraints in non-square weight matrices. IDInit focuses on initialization quality and does not address batch ordering or GPU determinism.

\subsection{ECG classification benchmarks}
The PTB-XL dataset~\cite{wagner2020ptbxl} and its associated benchmarks~\cite{strodthoff2021ptbxl} established xresnet1d101 (0.957 macro AUC, rhythm classification) as a widely used reference. Mehari and Strodthoff~\cite{mehari2022selfsupervised} demonstrated self-supervised pre-training for learning ECG representations from unlabeled data, achieving near-supervised performance with limited labels. Hannun et al.~\cite{hannun2019cardiologist} demonstrated cardiologist-level arrhythmia detection on ambulatory ECGs, and Ribeiro et al.~\cite{ribeiro2020automatic} achieved expert-level 12-lead ECG classification at scale, motivating potential clinical applications. Li et al.~\cite{li2025ecgfounder} trained ECGFounder on over 10 million ECGs with 150 label categories, demonstrating large-scale foundation model approaches to ECG analysis. None of these works address initialization-induced irreproducibility or provide deterministic training guarantees.

\subsection{Reproducibility in clinical AI}
Semmelrock et al.~\cite{semmelrock2025reproducibility} surveyed reproducibility barriers in ML-based research, finding that ``many papers are not even reproducible in principle'' due to sensitivity of training conditions. Desai et al.~\cite{desai2025reproducibility} proposed a taxonomy distinguishing repeatability, reproducibility, and replicability in AI/ML research. McDermott et al.~\cite{mcdermott2021reproducibility} found that ML for health consistently lags in reproducibility, with fewer than half of surveyed papers providing code. Pineau et al.~\cite{pineau2021improving} introduced the ML Reproducibility Checklist, raising code availability but not addressing algorithmic non-determinism. Bouthillier et al.~\cite{bouthillier2021accounting} demonstrated that ``variance due to data sampling, parameter initialization and hyperparameter choice impact markedly the results'' in ML benchmarks, motivating proper variance accounting. Zhuang et al.~\cite{zhuang2022randomness} characterized the impact of tooling randomness on neural network training, finding that while top-line metrics are minimally affected, performance on specific data subgroups is highly sensitive, consistent with the per-class instability we observe. Chen et al.~\cite{chen2022towards} proposed systematic frameworks for training reproducible deep learning models, identifying weight initialization as one of several reproducibility-critical random sources. Our approach eliminates this source at the algorithmic level.

Xie et al.~\cite{guo2025repdl} introduced RepDL, a library ensuring bit-level reproducibility across computing environments. RepDL addresses hardware-level non-determinism (floating-point associativity, platform differences). Our work addresses algorithmic non-determinism (initialization, batch ordering, GPU operations). These approaches are complementary.

\subsection{Neural collapse and classifier geometry}
Papyan et al.~\cite{papyan2020neural} discovered that during terminal training, classifier weights converge to a simplex Equiangular Tight Frame (ETF). Zhu et al.~\cite{zhu2021geometric} proved that ETF geometry is the unique global minimizer under cross-entropy loss with weight decay. Yang et al.~\cite{yang2022inducing} showed that ``feature learning with a fixed ETF classifier naturally leads to the neural collapse state even when the dataset is imbalanced among classes,'' motivating ETF initialization for imbalanced classification tasks. We exploit this by initializing classification heads from the ETF directly, providing structured geometry from the start (used heuristically in our multi-label sigmoid setting; see Methods).

\section{Methods}\label{sec:methods}

\subsection{Architectures}\label{sec:architectures}

\subsubsection{Conformer}
\begin{figure}[!t]
\centering
\includegraphics[width=0.82\textwidth]{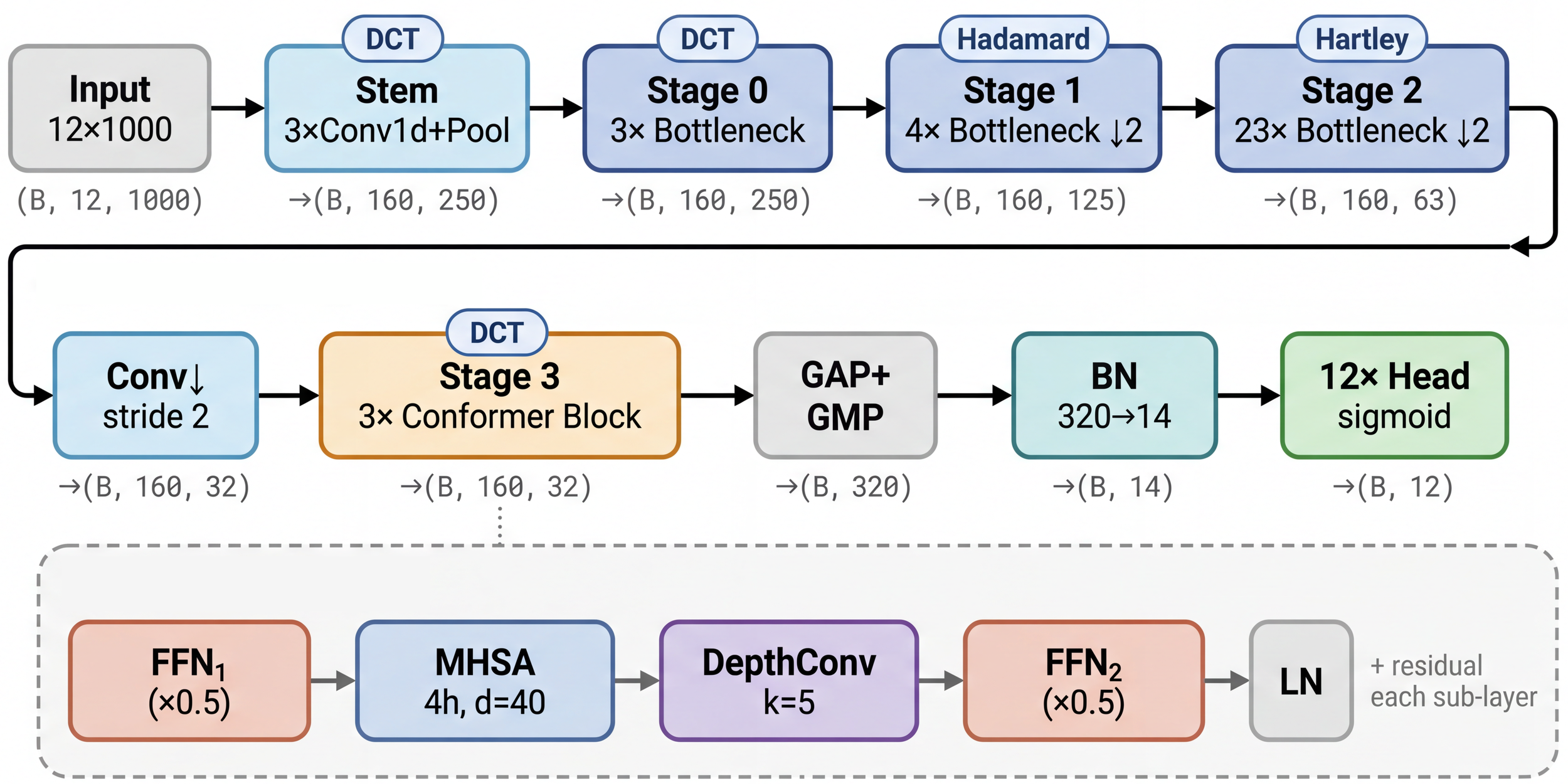}
\caption{ECG Conformer architecture (1.83M parameters). The network processes 12-lead ECG input (1000 samples at 100\,Hz) through a convolutional stem (3$\times$Conv1d + AvgPool, outputting 160 channels), three bottleneck stages with progressive stride-2 downsampling (3, 4, and 23 blocks respectively), a stride-2 transition convolution, and three Conformer blocks. Each Conformer block follows a macaron structure: FFN$_1$ (feed-forward network, $\times$0.5 residual) $\to$ MHSA (multi-head self-attention, 4 heads, $d\!=\!40$) $\to$ depthwise convolution ($k\!=\!5$) $\to$ FFN$_2$ ($\times$0.5 residual) $\to$ LN (layer normalization), with residual connections around each sub-layer. GAP (global average pooling) and GMP (global max pooling) are concatenated (320-d) and projected through a BN ($K\!+\!2$ information bottleneck, 14-d) before 12 independent sigmoid heads. Pill badges above each stage indicate the orthogonal basis used for weight initialization in mixed-basis mode. Color coding: light cyan = stem/transition layers, light blue = convolutional bottleneck stages, light orange = Conformer blocks, light teal = information bottleneck, light green = classification heads.}\label{fig:architecture}
\end{figure}

The Conformer (1.83M parameters), adapted from the speech recognition Conformer~\cite{gulati2020conformer}, combines depthwise separable convolutions with multi-head self-attention in a feed-forward--attention--convolution--feed-forward (FACF) block structure (Fig.~\ref{fig:architecture}). Input 12-lead ECGs (1000 time steps at 100\,Hz) are processed through a convolutional stem, followed by Conformer blocks that alternate global self-attention (capturing inter-beat temporal dependencies) with depthwise convolution (extracting local morphological features). The architecture uses 14-dimensional output features. Stride-2 skip connections in stages 1--2 use a deterministic adaptive average pooling implementation (custom autograd with explicit gradient distribution, avoiding non-deterministic atomicAdd). Conformer blocks (stage 3) contain no pooling operations. All CUDA operations are fully deterministic. Classification uses 12 independent sigmoid heads.

\subsubsection{Baseline CNN}
The baseline (1.65M parameters) follows an xresnet-style architecture~\cite{strodthoff2021ptbxl,howard2020fastai} with residual blocks at constant width, similar to the published xresnet1d101 but substantially smaller. It uses a 128-dimensional feature space with no intermediate bottleneck. Skip connections use adaptive average pooling for stride changes. Classification uses 12 independent sigmoid heads.

\subsection{Structured Orthogonal Initialization Algorithm}\label{sec:algorithm}

\subsubsection{Orthogonal basis computation}
For each convolutional layer Conv1d($C_\text{in}$, $C_\text{out}$, $K$), compute the DCT-II matrix $\mathbf{D}$ of size $C_\text{out} \times (C_\text{in} \cdot K)$. Each row is defined as:
\begin{equation}
d_i[j] = \cos\!\left(\frac{\pi \, i \, (2j + 1)}{2 \, C_\text{in} \, K}\right), \quad i \in [0, C_\text{out}),\ j \in [0, C_\text{in} \cdot K).\label{eq:dct}
\end{equation}
The resulting filters are near-orthogonal (worst-case maximum pairwise cosine similarity 0.138 for the stem layer where $C_\text{out} > \text{fan}_\text{in}$; most internal layers achieve $< 0.02$ after global mean subtraction).
For 2D convolutions, the basis uses a separable construction: 1D DCT along input channels combined with 2D DCT spatial basis functions over the kernel, yielding $n_\text{ch} \times n_\text{spatial}$ unique basis vectors. In bottleneck expand layers ($1\times1$ Conv2d: $d \to C_\text{out}$, $C_\text{out} \gg d$), only $d$ unique basis vectors exist ($n_\text{spatial} = 1$). Excess filters cycle through the basis modulo~$d$ (e.g., with $d\!=\!14$ and $C_\text{out}\!=\!512$, only 2.7\% of filters are unique). Despite this low uniqueness ratio, variance matching ensures correct activation magnitudes, and gradient descent rapidly diversifies the cycled filters during training (see MedMNIST results in Table~\ref{tab:medmnist}).

\subsubsection{Fixup residual scaling}
The last convolution in each residual branch (conv3) is scaled by $\alpha_{\mathrm{fixup}} = 0.01$: $\mathbf{w}_{\mathrm{conv3}} = 0.01 \cdot \mathbf{D}$, ensuring each residual block starts as near-identity ($y = x + \alpha \cdot f(x)$), providing gradient stability through deep networks.

For orthonormal initialization ($\kappa(\mathbf{W}) = 1$), each residual branch preserves input norms: $\|f_l(\mathbf{x})\| \leq \|\mathbf{x}\|$. The forward pass after $L$ blocks satisfies $\|\mathbf{y}_L\| \leq (1+\alpha)^L \|\mathbf{x}_0\| \approx e^{\alpha L}\|\mathbf{x}_0\|$, which remains bounded when $\alpha L = O(1)$, i.e., $\alpha = O(1/L)$. For $L=33$ blocks, this gives $\alpha \approx 0.03$. In contrast, Zhang et al.~\cite{zhang2019fixup} derive $\alpha = L^{-1/(2m-2)}$ for random initialization, where the spectral norm exceeds~1 due to the Marchenko--Pastur distribution. For $L=33$ and $m=3$ (bottleneck blocks), this gives $\alpha \approx 0.42$. The key distinction is that orthonormal matrices preserve norms exactly ($\kappa = 1$), making residual accumulation multiplicative rather than variance-additive, and requiring substantially smaller scaling.

The empirically optimal $\alpha = 0.01$ is consistent with $O(1/L)$ scaling. The additional reduction below~0.03 is attributable to GELU activation contraction ($\mathbb{E}[\mathrm{GELU}(x)^2] < \mathbb{E}[x^2]$), which we do not formally quantify. Ablation confirms the predicted ordering: $\alpha = 0$ is too conservative (0.946 AUC), $\alpha = 0.01$ is empirically optimal (0.961), the Fixup-derived $\alpha = 0.42$ is too aggressive for orthonormal weights (0.928), and $\alpha = 1.0$ causes gradient explosion (0.462).

\subsubsection{Variance matching}
All weights are zero-meaned and scaled to a target standard deviation:
\begin{equation}
\sigma = \frac{1}{\sqrt{3 \times \mathrm{fan}_{\mathrm{in}}}}\label{eq:sigma}
\end{equation}
This gives variance $1/(3 \cdot \mathrm{fan}_\mathrm{in})$, approximately $0.17\times$ the PyTorch \texttt{kaiming\_uniform\_} default ($2/\mathrm{fan}_\mathrm{in}$). The reduced scale was selected empirically; orthonormal initialization ($\kappa{=}1$) is less sensitive to the absolute variance scale than random initialization, because gradient flow is well-conditioned regardless of magnitude, and the fixup residual scaling ($\alpha{=}0.01$) compensates for the smaller activation magnitudes.

\subsubsection{ETF head geometry}
Classification head weights (12 heads in 14-dimensional space) are initialized from the Equiangular Tight Frame (ETF), the unique configuration of $K$ unit vectors in $\mathbb{R}^D$ ($D \geq K$) with minimum pairwise cosine similarity $-1/(K\!-\!1)$~\cite{papyan2020neural,zhu2021geometric}. The ETF is computed via SVD of the centered identity matrix:
\begin{equation}
\mathbf{M}_{\text{ETF}} = \text{SVD}\!\left(\mathbf{I}_K - \frac{1}{K}\mathbf{1}\mathbf{1}^\top\right).\label{eq:etf}
\end{equation}
ETF geometry is theoretically motivated by neural collapse under balanced softmax cross-entropy with weight decay~\cite{zhu2021geometric}. Li et al.~\cite{li2024neural} recently extended neural collapse theory to multi-label settings with pick-all-label loss, demonstrating that ETF geometry emerges as an optimal classifier structure even when multiple labels are active simultaneously. Their formulation uses summed softmax cross-entropies rather than our independent sigmoid BCE, so the ETF is used heuristically as a structured initialization rather than as a convergence target. The 14-dimensional bottleneck ($K\!+\!2 = 12\!+\!2$) provides the minimum representation capacity for $K$ classes plus a margin for inter-class separation, following the principle that useful representations require at least $K$ dimensions for $K$-class discrimination plus additional dimensions for features not captured by the class structure (e.g., OOD detection, calibration). The $K\!+\!2$ choice was validated empirically; other dimensionalities (e.g., $K\!+\!1$ or $K\!+\!3$) were not systematically compared.

\subsubsection{Numerical consistency}
DCT and sinusoidal basis matrices are computed in float64 precision, cached, and converted to float32, eliminating floating-point unit state leakage that could cause platform-dependent initialization differences. Hadamard matrices (binary $\pm 1$ entries) are exact in any precision; Hartley matrices are verified identical across tested platforms.

\subsubsection{Mixed orthogonal bases}
Any orthonormal basis can replace DCT in the initialization algorithm above. Our framework supports four bases: DCT-II (smooth cosine patterns), Hadamard (binary $\pm$1 patterns), Hartley (real-valued frequency decomposition), and sinusoidal (DST-II, sine patterns). Each basis can be used uniformly across all stages, or different bases can be assigned to different stages in a \emph{mixed} configuration.

The default mixed configuration assigns DCT-II to early stages, Hadamard to middle stages, Hartley to late stages, and DCT-II to the final stage. This assignment was selected heuristically by matching basis spectral properties to expected feature complexity at each depth, validated by ablation (Table~\ref{tab:ablation}), not derived from theory. One motivation is inter-stage decorrelation: for two orthonormal bases $\mathcal{B}_1, \mathcal{B}_2$ in $\mathbb{R}^n$, the mutual coherence $\mu(\mathcal{B}_1, \mathcal{B}_2) = \max_{i,j} |\langle \mathbf{b}_{1,i}, \mathbf{b}_{2,j}\rangle|$ satisfies $\mu < 1$ when the bases differ, potentially reducing cross-stage gradient coupling.

However, the basis comparison experiment at $n{=}20$ (Section~\ref{sec:basis}) shows that no structured basis significantly outperforms any other (Friedman $p{=}0.48$): all four single-basis configurations match Kaiming, with means ranging from 0.955 (Hadamard) to 0.959 (Hartley). The mixed configuration's variance reduction (std from 0.012 to 0.003; Table~\ref{tab:ablation}) may reflect the inclusion of low-variance bases (Hartley std=0.006, sinusoidal std=0.005) in the largest stage (23 of 30 bottleneck blocks) rather than inter-stage decorrelation per se. Mixed-basis remains the recommended default as the most thoroughly validated configuration. This design builds on ZerO initialization's~\cite{zhao2022zero} demonstration that Hadamard bases suffice for effective initialization.

Theoretical analysis of DCT initialization properties (energy preservation, condition number optimality, and connection to optimal decorrelation for first-order Markov processes) is provided in Appendix~\ref{app:theory}.

\subsection{Training Protocol}\label{sec:training}

All experiments use PyTorch 2.12 (nightly, CUDA 13.0) with CUDA deterministic execution. All ECG models are trained with Adam (lr=$10^{-3}$) and cosine annealing LR decay~\cite{loshchilov2017sgdr} over 85 epochs, batch size 128. Image classification experiments (CIFAR-100, MedMNIST) use SGD (lr=$0.1$, momentum $0.9$, weight decay $5 \times 10^{-4}$) with cosine annealing over 200 epochs. Binary cross-entropy loss is used for multi-label ECG classification. The sqrt class-weighted variant scales the positive-class weight for class~$k$:
\begin{equation}
w_k^{+} = \sqrt{\frac{N}{N_k}}\label{eq:classweight}
\end{equation}
where $N$ is the total number of training samples and $N_k$ is the number of positive training samples for class~$k$. No morphing or activation annealing is used. Buffer dimensions beyond $K$ classes are L2-regularized (weight 0.01). Logits are clamped to $[-50, 50]$ before loss computation for numerical stability. Validation AUC is evaluated every 10 epochs and at the final epoch; the checkpoint with highest validation AUC is selected for test evaluation. No-label samples (3.4\% of training data) are excluded. Data augmentation (random cropping, flipping) is not a source of non-determinism in this framework, as augmentation operations are deterministic given a fixed random generator state seeded per epoch.

Batch ordering uses one of two strategies. \emph{Seeded shuffle} (\texttt{torch.Generator} with a specified seed) is seed-dependent but data-independent: the permutation depends on the random seed, not on sample content, and changing the seed produces a different training run. \emph{Golden ratio ordering} is seed-free but data-dependent: the permutation is computed from signal content via
\begin{equation}
\text{key}[i] = \bigl(\text{hash}(X[i]) + \text{epoch} \cdot \varphi\bigr) \bmod 1\label{eq:golden}
\end{equation}
where $\text{hash}(X[i]) = \bigl(\sum|X[i]| \cdot \varphi\bigr) \bmod 1$ and $\varphi = (\sqrt{5}-1)/2$ is the golden ratio conjugate. Because the ordering depends only on the data and epoch index, it is fully deterministic with no random seed; the same dataset always produces the same batch sequence. Approximately 4\% of training records share float32 L1 norms (collision groups of size~$\leq 4$), producing fixed relative orderings for those pairs; permutation entropy loss is~$< 0.2\%$. The $\varphi$-rotation exploits equidistribution properties of irrational rotations~\cite{kuipers1974uniform}, producing a different but equally-spaced permutation each epoch. Low-discrepancy sequences have been shown to improve deep network training through more uniform input-space coverage~\cite{mishra2021lowdiscrepancy}; Bengio et al.~\cite{bengio2009curriculum} showed that presentation order affects learning outcomes. Deterministic GPU execution uses \texttt{torch.use\_deterministic\_algorithms(True)} with the environment variable \texttt{CUBLAS\_WORKSPACE\_CONFIG} set to \texttt{:4096:8}~\cite{paszke2019pytorch}.

\subsection{Deterministic Training Verification}\label{sec:verification}

For the Conformer, stride-2 skip connections use a custom deterministic adaptive average pooling implementation (explicit gradient distribution via \texttt{autograd.Function}, avoiding non-deterministic atomicAdd). Conformer blocks (stage 3) contain no pooling, so all operations are fully deterministic. For the baseline CNN, the \texttt{adaptive\_avg\_pool1d} backward pass uses non-deterministic atomicAdd in stride-2 skip connections. A custom \texttt{autograd.Function} replaces the backward pass with an explicit deterministic loop, adding approximately 1\% training overhead (compared to 20--30\% for global \texttt{torch.use\_deterministic\_algorithms}~\cite{shanmugavelu2024impacts}). Bit-identical training is verified by MD5 hash comparison of trained model parameters across independent runs on the same hardware.

The 2D CIFAR experiments validate initialization quality under standard training conditions (cuDNN auto-tuning enabled), comparing DCT vs.\ Kaiming with 20 seeds per condition. Full determinism is demonstrated on MedMNIST (below). For MedMNIST, we enable full determinism: deterministic algorithms, deterministic cuBLAS workspace, disabled cuDNN benchmarking, single-threaded data loading, and per-epoch seeded augmentation. Under these conditions, the DCT golden configuration produces bit-identical trained models, verified by running each experiment twice and comparing MD5 hashes of all model parameters. Three representative datasets confirmed: PathMNIST (\texttt{62e7d8\allowbreak{}6bfb32\allowbreak{}55c319\allowbreak{}f0efbb\allowbreak{}f95492cc}), DermaMNIST (\texttt{eff682\allowbreak{}38d9a7\allowbreak{}91df4e\allowbreak{}628ed9\allowbreak{}6893be8c}), BloodMNIST (\texttt{e06346\allowbreak{}1547d3\allowbreak{}8d573c\allowbreak{}b7050b\allowbreak{}8f05a4ec}). The DCT initialization itself is fully deterministic: the same MD5 hash (\texttt{c64e66\allowbreak{}e8764d\allowbreak{}8b749e\allowbreak{}b794b7\allowbreak{}9b11638c}) is produced across NVIDIA RTX~3080, RTX~5090, A100, and RTX Pro~6000 GPUs.

\subsection{Methodological Note: Multiple Comparisons}\label{sec:methodo_note}

Approximately 140 configurations were evaluated on fold~10 during model development. The fold-10 results should be interpreted as development metrics, not unbiased estimates. Our primary claims are validated on genuinely held-out data: (1)~cross-dataset generalization on three external ECG databases (AFDB, CPSC2018, Chapman-Shaoxing) never used during model development (Section~\ref{sec:crossdataset}), (2)~cross-domain validation on seven MedMNIST medical image benchmarks with $n{=}20$ seeds per condition (Section~\ref{sec:medmnist}), and (3)~CIFAR-100 equivalence testing with $n\!=\!20$ seeds (Appendix~\ref{app:cifar}). The MedMNIST and CIFAR-100 experiments were conducted after all ECG architecture development and hyperparameter selection were complete. These are confirmatory experiments on new domains and datasets, not additional comparisons within the development loop, and are therefore not subject to the multiple comparisons concern. Note: fold~9 was used for validation during architecture selection. Its inclusion as one of three CV test folds may slightly inflate the CV mean; folds~8 and~10 are uncontaminated.
\subsection{Statistical Power Analysis}\label{sec:power}

Table~\ref{tab:power} reports the statistical power available for each primary comparison. Both ECG comparisons reach statistical significance at $n\!=\!20$: Conformer ($p = 0.016$, 71\% power) and Baseline CNN ($p < 0.001$, $>$99\% power). The paper's primary claim is equivalence of structured initialization to Kaiming. TOST equivalence is confirmed for the Conformer ($p < 0.001$ at $\delta\!=\!0.015$, chosen as 1.5$\times$ the Kaiming within-condition standard deviation, representing a difference smaller than natural seed-to-seed variation) and on CIFAR-100 ($p < 0.001$ at $\delta\!=\!0.5$ percentage points, $n\!=\!20$; Appendix~\ref{app:cifar}). Both ECG comparisons additionally reach significance for superiority: the Conformer advantage (0.6 percentage points) is clinically negligible for macro AUC, while the Baseline advantage (0.9 percentage points, $d = 1.60$) reflects a large effect driven by Kaiming's $2.8{\times}$ higher variance.

\begin{table}[!t]
\caption{Statistical power analysis for primary comparisons. Observed Cohen's $d$ computed from pooled standard deviation; power estimated for two-sided independent $t$-test at $\alpha\!=\!0.05$; $n_{80\%}$ is the per-group sample size required for 80\% power.}\label{tab:power}
\centering
\footnotesize
\begin{tabular}{@{}lcccrl@{}}
\toprule
Comparison & $n$ & $d$ & Power & $n_{80\%}$ & TOST \\
\midrule
Conform.\ mixed vs Kaim. & 20 & 0.82 & $\sim$71\% & 25 & $p < 0.001$ ($\delta\!=\!0.015$) \\
Baseline struct.\ vs Kaim. & 20 & 1.60 & $>$99\% & 7 & superiority ($p < 0.001$) \\
CIFAR-100 struct.\ vs Kaim. & 20 & 0.24 & 11\%\textsuperscript{a} & --- & $p < 0.001$ ($\delta\!=\!0.5$pp) \\
\bottomrule
\end{tabular}
\par\smallskip
\footnotesize{\textsuperscript{a}Low superiority power is expected: the observed difference is small ($d\!=\!0.24$) and the primary claim is equivalence (TOST $p < 0.001$).}
\end{table}

\subsection{Data}\label{sec:data}

\subsubsection{PTB-XL}
PTB-XL v1.0.3~\cite{wagner2020ptbxl}: 21,799 12-lead ECG recordings at 100\,Hz, 10 seconds each. We classify 12 rhythm superclasses (SR, AFIB, STACH, SARRH, SBRAD, PACE, SVARR, BIGU, AFLT, SVTAC, PSVT, TRIGU) as a multi-label task. Class distribution is severely imbalanced: SR comprises 77.0\% of training samples; six classes have fewer than 200 training samples (PACE: 237, SVARR: 128, BIGU: 66, AFLT: 59, SVTAC: 21, PSVT: 19, TRIGU: 16). Test-set (fold~10) positive counts per class: SR: 1674, AFIB: 152, STACH: 82, SARRH: 77, SBRAD: 64, PACE: 28, SVARR: 14, BIGU: 8, AFLT: 7, SVTAC: 3, PSVT: 2, TRIGU: 2. Approximately 96\% of records have a single rhythm label, and 3.4\% have no rhythm label and are excluded from training. Data split follows~\cite{wagner2020ptbxl,strodthoff2021ptbxl}: strat\_fold 1--8 train (17,418 records), fold 9 validation (2,183), fold 10 test (2,198). Z-normalization uses train-set statistics only. For 6-lead cross-dataset experiments, we use leads I, II, III, aVR, aVL, aVF.

\subsubsection{MIT-BIH AFDB}
MIT-BIH AFDB~\cite{moody1983afib}: 25 long-term (10-hour) two-channel Holter recordings with beat-level AFIB annotations, originally sampled at 250\,Hz. We anti-alias downsample to 100\,Hz using a polyphase FIR filter (Kaiser window, applied via \texttt{scipy.signal.resample\_poly}, deterministic, no data-dependent parameters) and extract 84,334 non-overlapping 10-second windows. The two available channels are mapped to leads I and II, with leads III, aVR, aVL, and aVF derived via Einthoven's law and Goldberger's equations, providing 6 limb leads but no precordial information (a distribution shift from the 6-lead PTB-XL training data). No training is performed on AFDB; it is used exclusively for zero-shot cross-dataset evaluation of AFIB detection.

\subsubsection{CPSC2018 and Chapman-Shaoxing}
The China Physiological Signal Challenge 2018 dataset (CPSC2018)~\cite{cpsc2018} (500\,Hz, 12-lead) and the Chapman-Shaoxing 12-lead ECG database~\cite{zheng2020chapman} (500\,Hz, 12-lead) provide additional external validation. Both are anti-alias downsampled to 100\,Hz using the same polyphase FIR filter as AFDB. Both datasets contain AFIB annotations and were never used during model development, architecture selection, or hyperparameter tuning. For 6-lead evaluation, we use the same lead subset as PTB-XL (I, II, III, aVR, aVL, aVF). All external datasets are normalized using PTB-XL training-set statistics only, preventing information leakage.

\subsubsection{MedMNIST}
Seven datasets from the MedMNIST v2 benchmark~\cite{yang2023medmnist} provide cross-domain medical image validation: BloodMNIST (17,092 blood cell images, 8 types), OrganCMNIST (23,583 abdominal CT slices, 11 organ classes), DermaMNIST (10,015 dermatoscopic images, 7 lesion categories), BreastMNIST (780 breast ultrasound images, 2 classes), RetinaMNIST (1,600 fundus images, 5 ordinal grades), ChestMNIST (112,120 chest X-rays, 14 multi-label findings), and PathMNIST (107,180 colon pathology patches, 9 tissue types). All images are $28 \times 28$ RGB. We train ResNet-18~\cite{he2016deep} with a modified stem (3$\times$3 convolution stride~1, no initial max pooling) without an information bottleneck. Training uses SGD (lr=$0.1$, momentum $0.9$, weight decay $5 \times 10^{-4}$) with cosine annealing over 200 epochs and standard augmentation (random crop, horizontal flip). Each condition (DCT, Kaiming) uses 20 seeds plus one golden ratio deterministic run, totaling 42 runs per dataset (294 total).

\subsection{Evaluation Metrics}\label{sec:metrics}

Primary metric: macro AUC (area under receiver operating characteristic curve, averaged across 12 classes). All cross-validation standard deviations use the sample standard deviation (dividing by $n-1$). Multi-seed standard deviations also use $n-1$. Secondary metrics: effective rank (eRank via SVD entropy of feature matrix) and expected calibration error (ECE, 10 bins per class). Cross-validation uses 3 folds (test on folds 10, 9, 8) with the standard training protocol.

\section{Results}\label{sec:results}

\subsection{Structured Initialization and the Combined Framework}\label{sec:framework}

The structured initialization framework (Section~\ref{sec:algorithm}) replaces random weight sampling with analytically computed orthogonal basis functions, scaled to match Kaiming variance, with fixup residual scaling ($\alpha = 0.01$), mixed orthogonal bases across stages (for ECG), and ETF-initialized classification heads.

Single-basis DCT initialization matches Kaiming in mean performance ($0.956$ vs $0.953$ at $n{=}20$; Tables~\ref{tab:main},~\ref{tab:basis}) but shows higher batch-ordering sensitivity. Table~\ref{tab:ablation} presents the component ablation.

\begin{table}[!t]
\caption{Ablation of structured initialization framework components (Conformer architecture). ``DCT-only'' = single-basis DCT across all stages, no class weighting. ``Mixed'' = DCT/Hadamard/Hartley per stage. ``cw'' = sqrt class weighting. Standard deviations reflect batch-ordering variance ($n\!=\!5$ batch seeds per configuration); the Kaiming value differs from Table~\ref{tab:main} ($n\!=\!20$ initialization seeds) because it measures a different source of variance.}\label{tab:ablation}
\centering
\begin{tabular}{lll}
\toprule
Component & Test AUC & Delta \\
\midrule
DCT-only (fixup=0.01) & $0.948 \pm 0.012$ & baseline \\
+ mixed bases & $0.951 \pm 0.006$ & +0.003 \\
+ sqrt class weight & $0.951 \pm 0.006$ & +0.003 \\
+ mixed + cw (``Mixed'') & $0.961 \pm 0.003$ & +0.013 \\
Kaiming (reference) & $0.958 \pm 0.002$ & --- \\
Kaiming + cw & $0.960 \pm 0.006$\textsuperscript{a} & +0.002 \\
\bottomrule
\end{tabular}
\par\smallskip
\footnotesize{\textsuperscript{a}Kaiming+cw: $n{=}3$ converged of 5 launched; 2 seeds collapsed to $<$0.60 AUC (0\% failure rate for all structured configurations). DCT-only batch-ordering mean (0.948) reflects variance at $n{=}5$ batch seeds with fixed init; at $n{=}20$ init seeds (Table~\ref{tab:basis}), DCT matches Kaiming (0.956 vs 0.953).}
\end{table}

Plain DCT initialization produces higher batch-ordering sensitivity (std=0.012 vs Kaiming's 0.002) because all DCT filters are smooth cosines that produce correlated gradients within each stage. Mixed orthogonal bases introduce gradient diversity across stages, and sqrt class weighting amplifies signal from rare classes. Neither component individually exceeds Kaiming ($0.951 \pm 0.006$ for each, vs Kaiming $0.958 \pm 0.002$), and their combination (0.961) numerically exceeds Kaiming (0.958) within the batch-ordering variance analysis ($n\!=\!5$). At $n{=}20$ init seeds (Table~\ref{tab:basis}), all single-basis structured initializations match Kaiming, confirming that batch-ordering sensitivity (measured here) is distinct from initialization quality. At $n{=}20$, sqrt class weighting does not improve Kaiming ($0.960 \pm 0.006$, $n{=}3$ converged of 5, vs $0.953 \pm 0.010$ without cw) and 2 of 5 seeds collapsed to $<$0.60 AUC (consistent with the $n{=}20$ result where 2 of 20 seeds collapsed). Throughout this paper, ECG experiments labeled ``Mixed'' in tables use the combined framework (mixed-basis stages + sqrt class weighting); MedMNIST and CIFAR experiments use single-basis DCT. ``Structured initialization'' refers to any deterministic orthogonal basis (DCT, Hadamard, Hartley, or sinusoidal), in contrast to random Kaiming initialization.

\subsection{Conformer Multi-Seed Validation}\label{sec:conformer}

\begin{figure}[!t]
\centering
\includegraphics[width=\linewidth]{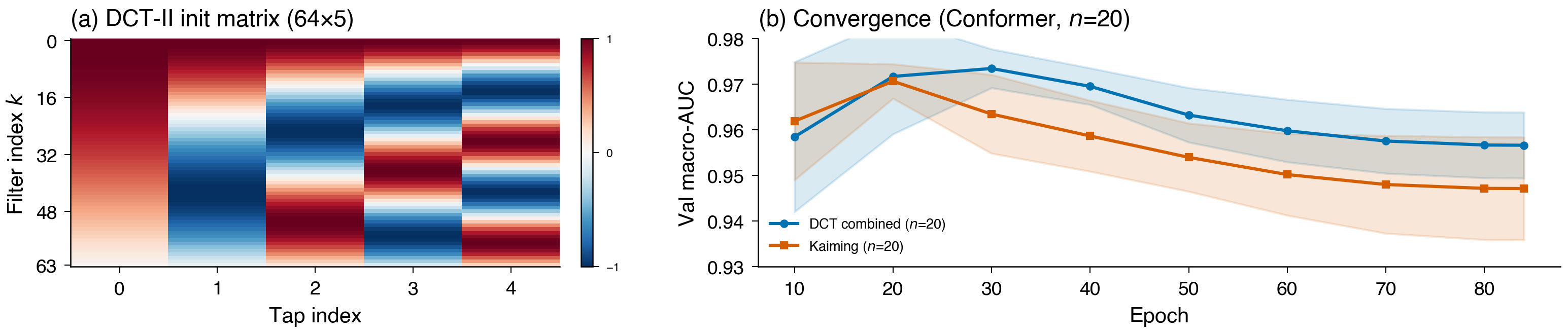}
\caption{Structured orthogonal initialization. (a)~DCT-II basis matrix for a Conv1d(64, 64, 5) layer: all 64 cosine basis vectors spanning from DC ($k\!=\!0$, top) to near-Nyquist ($k\!=\!63$, bottom). Each row is one deterministic filter; the structured frequency progression replaces random Kaiming weights. Alternative bases (Hadamard, Hartley, sinusoidal) use analogous constructions. (b)~Convergence comparison (Conformer, $n\!=\!20$ seeds): mean $\pm$1 std bands. Mixed-basis initialization maintains higher validation AUC with lower variance from epoch~20 onward. Test-set results in Table~\ref{tab:main}.}\label{fig:method}
\end{figure}

Mixed-basis initialization was applied to a Conformer architecture (1.83M parameters) combining depthwise separable convolutions with multi-head self-attention~\cite{gulati2020conformer} (Fig.~\ref{fig:method}). The Conformer, adapted from speech recognition, uses a custom deterministic adaptive average pooling implementation in its stride-2 skip connections (stages 1--2), while Conformer blocks (stage 3) contain no pooling operations. Combined with deterministic GPU execution mode, all CUDA operations are fully deterministic.

The Conformer with mixed-basis initialization yielded $0.959 \pm 0.005$ test macro AUC (20 batch seeds), exceeding the published xresnet1d101 benchmark ($0.957$, single seed, 9.0M parameters~\cite{strodthoff2021ptbxl}) with $5{\times}$ fewer parameters (1.83M). Our reproduction of xresnet1d101 with 5 Kaiming seeds yields $0.947 \pm 0.013$, 1.0 percentage point below the published result and $3{\times}$ the variance, illustrating the unreliability of single-seed reporting. Structured initialization is statistically significantly higher than Kaiming on the same Conformer architecture ($0.959 \pm 0.005$ vs $0.953 \pm 0.010$, $p = 0.016$, Welch $t$-test; Table~\ref{tab:main}). Kaiming initialization exhibits twice the variance of structured init (std 0.010 vs 0.005), with individual seeds ranging from 0.926 to 0.967. A fully deterministic variant (golden ratio batch ordering, zero random seeds at any stage, bit-identical model weights verified by MD5 hash) reached 0.966 on the standard evaluation fold (fold 10). Three-fold cross-validation with seeded batch ordering confirms stable performance ($0.955 \pm 0.012$; folds: 0.942, 0.961, 0.963). The fully deterministic configuration (golden ratio batching) shows higher fold variance ($0.941 \pm 0.031$), reflecting sensitivity to fold-level data composition rather than initialization.

The 0.966 single-fold golden result falls within the seeded CV fold range (0.942--0.963); the CV means of 0.941 (golden ratio) or 0.955 (seeded batch ordering) provide more representative estimates of expected performance.

With $n\!=\!20$ seeds per condition, the 95\% confidence interval for the mean difference is $[0.001, 0.011]$ AUC, excluding zero. Cohen's $d = 0.82$ (95\% CI $[0.15, 1.46]$), a large effect driven primarily by Kaiming's higher variance, not a large mean difference. Mixed-basis initialization is equivalent to Kaiming within the $\delta{=}0.015$ margin (TOST $p < 0.001$). The mean difference of 0.6 percentage points is statistically detectable ($p = 0.016$) but clinically negligible for macro AUC.

\begin{table}[!t]
\caption{Structured initialization across architectures (PTB-XL rhythm classification, 12-class multi-label, macro AUC). ``Det.'' indicates determinism level. ``Mixed'' = mixed-basis initialization (DCT/Hadamard/Hartley per stage); ``DCT'' = single-basis DCT initialization. Both use sqrt class weighting. MedMNIST cross-domain results in Table~\ref{tab:medmnist}; CIFAR-100 in Appendix~\ref{app:cifar}.}\label{tab:main}
\centering
\footnotesize
\begin{tabular}{@{}llllrl@{}}
\toprule
Model & Init & Configuration & Test AUC & Params & Det. \\
\midrule
\textbf{Conformer} & \textbf{Mixed} & \textbf{20 batch seeds} & $\mathbf{0.959 \pm 0.005}$ & \textbf{1.83M} & \textbf{Init} \\
Conformer & Mixed & golden ratio & 0.966\textsuperscript{a} & 1.83M & Full \\
Conformer & Kaiming & 20 init seeds & $0.953 \pm 0.010$ & 1.83M & No \\
Baseline CNN & DCT & 20 batch seeds & $0.956 \pm 0.004$ & 1.65M & Init \\
Baseline CNN & Kaiming & 20 init seeds & $0.947 \pm 0.007$ & 1.65M & No \\
xresnet1d101 & Kaiming & 5 init seeds & $0.947 \pm 0.013$ & 9.0M & No \\
xresnet1d101~\cite{strodthoff2021ptbxl} & Kaiming & published & 0.957 & 9.0M & No \\
\bottomrule
\end{tabular}
\vspace{2pt}

\footnotesize{\textsuperscript{a}Single fold~10. CV: $0.955 \pm 0.012$ (seeded batch ordering), $0.941 \pm 0.031$ (golden ratio batch ordering). Initialization is deterministic in both cases; the variance reflects batch ordering only.}
\end{table}

\subsection{Variance Decomposition}\label{sec:variance}

Training variance decomposes into four independently addressable components. \emph{Initialization variance} is zero by construction, as structured orthogonal basis functions produce identical weights on every run. \emph{Batch-ordering variance} measured std=0.005 under seeded shuffle ($n\!=\!20$) and zero under golden ratio scheduling. \emph{GPU operation variance} is zero for the Conformer, which uses deterministic adaptive pooling implementations; for the baseline CNN, non-deterministic atomicAdd in skip connections is addressed via custom autograd at 1\% training overhead. \emph{Fold/data composition variance} measured std=0.012 (3-fold CV, estimated from $n=3$) and cannot be eliminated as it reflects genuine data variation. The fold variance estimate depends on the batch ordering strategy: seeded shuffle yielded fold std=0.012 while golden ratio yielded std=0.031, indicating interaction between batch ordering and fold/data variance. The decomposition assumes approximate independence, and the sources do interact.

The Conformer with golden ratio batch ordering eliminates all three controllable sources, achieving MD5-verified bit-identical training across independent runs.

\subsection{Orthogonal Basis Comparison}\label{sec:basis}

To test whether the choice of orthogonal basis affects learned representations, we trained the Conformer with four structured initializations (DCT, Hartley, Hadamard, and sinusoidal), each across 20 initialization seeds on the PTB-XL rhythm classification task (Table~\ref{tab:basis}). Because the same 20 seeds are used across all bases (producing identical batch orderings), pairwise comparisons use paired $t$-tests. Kaiming+cw uses the same seed numbers, providing matched batch orderings.

\begin{table}[!t]
\centering
\caption{Orthogonal basis comparison (Conformer, PTB-XL rhythm classification, $n{=}20$ seeds each). Each row uses a \emph{single} basis uniformly across all stages with sqrt class weighting. Paired $t$-tests (same seeds across bases). Friedman $\chi^2{=}2.46$, $p{=}0.48$: no significant difference between structured bases. All standard deviations use $n{-}1$ (sample std). \textsuperscript{a}Kaiming+cw: $n{=}18$ converged of 20 launched; 2 seeds (2, 7) collapsed to $<$0.67 AUC (0\% failure rate for all structured bases and Kaiming without cw).}\label{tab:basis}
\begin{tabular}{lcc}
\toprule
Basis (single, all stages) & Test AUC (mean $\pm$ std) & vs.\ DCT $p$ (paired) \\
\midrule
Hartley-only    & $\mathbf{0.959 \pm 0.006}$ & $0.253$ \\
Sinusoidal-only & $0.956 \pm 0.005$          & $0.940$ \\
Hadamard-only   & $0.955 \pm 0.008$          & $0.751$ \\
DCT-only        & $0.956 \pm 0.010$          & --- \\
\midrule
Kaiming+cw      & $0.954 \pm 0.012$\textsuperscript{a} & --- \\
\bottomrule
\end{tabular}
\end{table}

At $n{=}20$, no structured basis significantly outperforms any other (Friedman $\chi^2{=}2.46$, $p{=}0.48$). All four bases match Kaiming (individual paired comparisons: all $p > 0.25$), with means ranging from $0.955$ (Hadamard) to $0.959$ (Hartley). No pairwise comparison survives Holm--Bonferroni correction. The result establishes that the contribution is deterministic structured initialization itself, not any particular basis function.

The key finding is a \emph{variance gradient} across bases: DCT shows the highest structured training variance (std=0.010), followed by Hadamard (0.008), Hartley (0.006), and sinusoidal (0.005), compared to Kaiming's 0.010 (Table~\ref{tab:main}). Levene's test shows no significant pairwise variance differences (all $p > 0.72$), but the monotonic ordering (smooth periodic bases producing the lowest variance, binary patterns intermediate, and purely cosine the highest) is consistent with basis smoothness governing optimization stability. A largely concordant ordering appears in noise robustness (Appendix Table~\ref{tab:noise}): Hartley $>$ Hadamard $>$ Sinusoidal $>$ DCT across all three noise types (Friedman $p{=}0.069$ for electrode motion; sinusoidal and Hartley swap positions relative to the variance ranking). All structured bases exhibit equal or lower variance than Kaiming.

Per-class analysis at $n{=}20$ shows one significant class-level effect: atrial flutter (AFLT: Friedman $p{=}0.046$), where Hartley leads ($0.917$) and Hadamard trails ($0.893$). The aggregate per-class pattern shows no single basis dominating: Hartley is best on 5 of 12 classes, Hadamard on 3, sinusoidal on 2, and DCT on 2.

\subsection{Golden Deterministic Baselines}\label{sec:golden}

To isolate the effect of initialization from batch ordering, we trained both architectures with golden ratio deterministic batch ordering using mixed-basis and Kaiming initialization (Table~\ref{tab:golden}).

\begin{table}[!t]
\centering
\caption{Golden-seed $2{\times}2$ design: architecture crossed with initialization, all using golden ratio batch ordering. ``Mixed'' = mixed-basis (DCT/Hadamard/Hartley per stage) with sqrt class weighting. Per-class AUC shown for the three rhythms with the largest initialization effect. Each entry is a single deterministic run (MD5-verified for structured init). Caution: TRIGU ($n\!=\!2$ test positives), SVARR ($n\!=\!14$), and AFLT ($n\!=\!7$) have small test-set sizes; per-class AUC estimates carry substantial uncertainty.}\label{tab:golden}
\begin{tabular}{llcccc}
\toprule
Architecture & Init & AUC & TRIGU & SVARR & AFLT \\
\midrule
Conformer & Mixed   & $\mathbf{0.959}$ & $\mathbf{0.980}$ & $0.892$ & $\mathbf{0.951}$ \\
Conformer & Kaiming & $0.928$          & $0.781$          & $0.923$ & $0.855$ \\
\midrule
Baseline  & Mixed   & $0.950$          & $0.906$          & $0.870$ & $0.963$ \\
Baseline  & Kaiming & $0.940$          & $0.891$          & $0.814$ & $0.941$ \\
\bottomrule
\end{tabular}
\end{table}

Mixed-basis initialization improves overall AUC in both the Conformer ($+3.1$ percentage points versus Kaiming) and the Baseline ($+1.0$\,pp). The Conformer--Kaiming combination illustrates a seed-dependent per-class failure on trigeminy (TRIGU AUC $= 0.781$ versus $0.980$ with structured init, a 19.9\,pp gap) and atrial flutter (AFLT: $0.855$ vs.\ $0.951$, 9.6\,pp). While TRIGU has only $n{=}2$ test positives in fold~10, this variability is independently confirmed by: (a)~3-fold CV, where each fold tests TRIGU on different samples ($0.885 \pm 0.167$ under golden Conformer), and (b)~the $n{=}20$ multi-seed distribution (TRIGU range: 30.9pp under Kaiming vs 4.1pp under structured init, Fig.~\ref{fig:perclass}). The Baseline--Kaiming model avoids such extreme failures (TRIGU $= 0.891$), indicating that the deeper Conformer is more sensitive to initialization quality for rare rhythms. Supraventricular arrhythmia (SVARR) shows a consistent pattern across architectures: Kaiming drops to $0.814$ in the Baseline versus $0.870$ with structured initialization.

\subsection{Cross-Domain Medical Image Validation}\label{sec:medmnist}

To validate structured initialization beyond 1D ECG signals, we applied single-basis 2D-DCT initialization to ResNet-18 on seven MedMNIST medical image benchmarks~\cite{yang2023medmnist} spanning seven clinical domains (Table~\ref{tab:medmnist}). MedMNIST uses single-basis DCT rather than mixed-basis because ResNet-18 lacks the multi-stage bottleneck structure that motivated mixed assignment in the ECG architectures; golden runs with alternative bases confirm that basis choice has minimal impact on 2D imaging tasks (see below). Each experiment uses 20 random seeds per initialization method (DCT and Kaiming) plus one golden ratio deterministic run per method, enabling proper statistical comparison.

\begin{table}[!t]
\caption{MedMNIST cross-domain validation. ResNet-18, 200 epochs, $n{=}20$ seeds per initialization. DCT = single-basis 2D-DCT (no mixed bases, no class weighting). $p$-values from Welch's $t$-test. No dataset shows a statistically significant difference.}\label{tab:medmnist}
\centering
\footnotesize
\begin{tabular}{@{}lccccr@{}}
\toprule
Dataset & DCT AUC & Kaiming AUC & $\Delta$ & $p$ & $n_\text{train}$ \\
\midrule
BloodMNIST   & $.9986\!\pm\!.0002$ & $.9987\!\pm\!.0002$ & $-.0001$ & .65 & 11{,}959 \\
OrganCMNIST  & $.9943\!\pm\!.0006$ & $.9945\!\pm\!.0004$ & $-.0002$ & .16 & 13{,}000 \\
DermaMNIST   & $.9186\!\pm\!.0062$ & $.9192\!\pm\!.0048$ & $-.0006$ & .74 & 7{,}007 \\
BreastMNIST  & $.878\!\pm\!.032$   & $.877\!\pm\!.030$   & $+.001$  & .96 & 546 \\
RetinaMNIST  & $.729\!\pm\!.016$   & $.734\!\pm\!.014$   & $-.005$  & .26 & 400 \\
\midrule
ChestMNIST   & $.756\!\pm\!.011$   & $.752\!\pm\!.014$   & $+.004$  & .32 & 78{,}468 \\
PathMNIST    & $.9903\!\pm\!.0034$ & $.9899\!\pm\!.0029$ & $+.000$  & .74 & 89{,}996 \\
\bottomrule
\end{tabular}
\end{table}

Across all seven datasets, no statistically significant difference exists between DCT and Kaiming initialization (all $p > 0.14$, Cohen's $d$ range: $-$0.46 to $+$0.32). On saturated benchmarks (BloodMNIST AUC~0.999, OrganCMNIST~0.994, PathMNIST~0.990), both methods reach near-ceiling performance. On datasets with more headroom (ChestMNIST $\sim$0.75 AUC, RetinaMNIST $\sim$0.73, BreastMNIST $\sim$0.88), the difference remains negligible. All DCT golden results are bit-identical across independent runs (MD5 verified). MedMNIST experiments use a fixed architecture and hyperparameter recipe transferred from the ECG pipeline without task-specific optimization. They validate that DCT initialization generalizes across modalities without degrading performance, not that it achieves state-of-the-art on each benchmark.

Golden ratio deterministic runs with alternative bases (Hadamard, Hartley, Sinusoidal) on DermaMNIST and RetinaMNIST confirm that basis choice has minimal impact on 2D imaging tasks: DermaMNIST golden AUCs cluster within 0.004 across all five bases (DCT 0.915, Hadamard 0.914, Hartley 0.911, Sinusoidal 0.914, Kaiming 0.912). On RetinaMNIST, DCT leads (0.755) but three of four structured bases fall below Kaiming (0.733): Hadamard (0.701), Hartley (0.711), and sinusoidal (0.717), consistent with the high-variance regime of this small dataset ($n_\text{train}\!=\!400$).

\begin{figure}[!t]
\centering
\includegraphics[width=\linewidth]{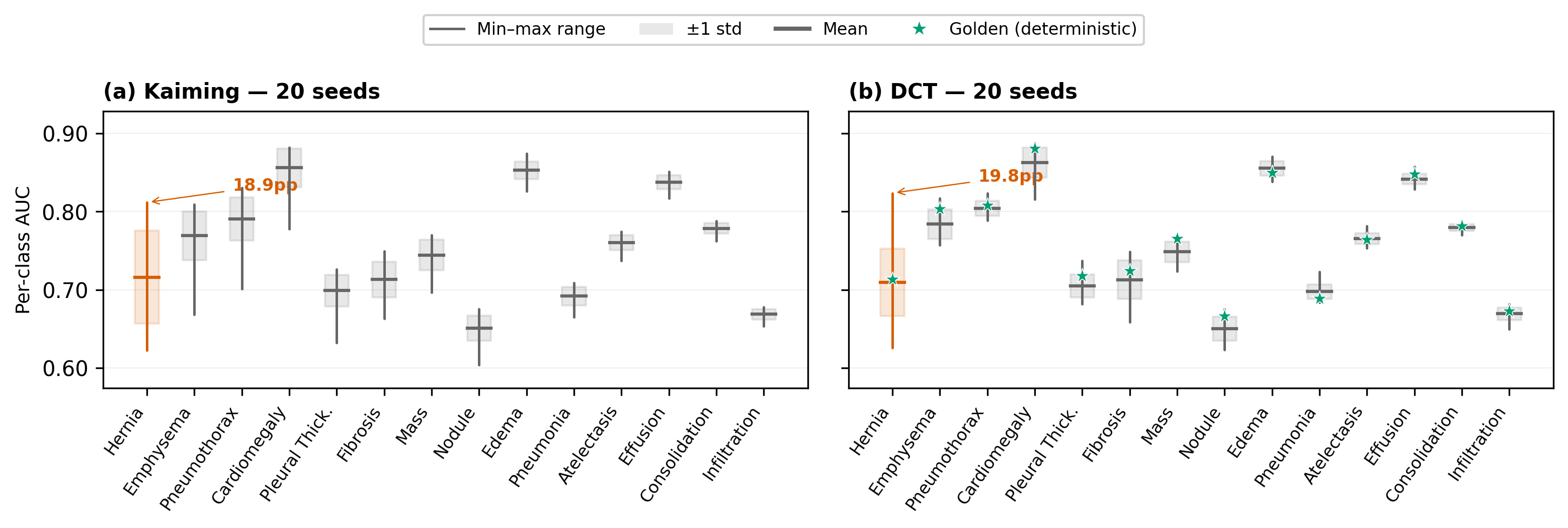}
\caption{Per-class AUC variability on ChestMNIST (ResNet-18, single-basis 2D-DCT, $n\!=\!20$ seeds). (a)~Kaiming initialization: Pneumothorax ($n_\text{pos}\!=\!1{,}089$) ranges 0.130, Emphysema ($n_\text{pos}\!=\!509$) ranges 0.141. (b)~DCT initialization: systematically tighter ranges in 11 of 14 classes (mean range 0.056 vs 0.079). Pneumothorax tightens $3.7{\times}$ (0.035 vs 0.130), Emphysema $2.4{\times}$. Green stars: DCT golden deterministic run. Thick horizontal lines: mean; shaded regions: $\pm$1 std; thin whiskers: full min--max range. Compare with Fig.~\ref{fig:perclass} for the analogous ECG pattern.}\label{fig:chestmnist}
\end{figure}

Per-class analysis on ChestMNIST (14 thorax diseases, multi-label, imbalanced) shows that the variance reduction observed in ECG generalizes to imbalanced imaging tasks (Fig.~\ref{fig:chestmnist}). DCT initialization produces a tighter per-class AUC range than Kaiming in 11 of 14 disease classes, with an overall mean per-class range of 0.056 (DCT) vs 0.079 (Kaiming). The strongest stability advantage appears on uncommon conditions: Pneumothorax (Kaiming range 3.7$\times$ DCT range, $n_\text{pos}\!=\!1{,}089$ test cases) and Emphysema (2.4$\times$, $n_\text{pos}\!=\!509$). Unlike the ECG rare classes where small test sets ($n_\text{pos}\!\leq\!14$) limit per-class AUC precision, ChestMNIST has sufficient test cases for reliable per-class estimates. The rarest class (Hernia, $n_\text{pos}\!=\!42$) exhibits high variance under both methods (range $\sim$0.19).

These aggregate results complement the ECG findings (Tables~\ref{tab:main},~\ref{tab:basis},~\ref{tab:golden}) where structured initialization provides significant advantages at the macro level. Together, the results demonstrate a consistent pattern across domains: on saturated benchmarks, initialization is irrelevant; on imbalanced clinical tasks, per-class variance reduction emerges (ChestMNIST: 11 of 14 classes, Pneumothorax $3.7{\times}$ reduction) and reaches aggregate statistical significance on specialist tasks (ECG: $p{=}0.016$). This pattern holds independently in 1D (ECG rhythm classification) and 2D (chest X-ray multi-label classification), confirming that the variance reduction is a property of structured initialization, not task-specific.

\subsection{Architecture Generality: Baseline CNN Matches Published SOTA}\label{sec:baseline}

To establish that structured initialization generalizes beyond the Conformer, we applied single-basis DCT initialization to a standard xresnet-style CNN (1.65M parameters, 128-dimensional feature space, no bottleneck, no self-attention). This architecture is structurally similar to the published xresnet1d101~\cite{strodthoff2021ptbxl} but smaller (1.65M vs 9.0M parameters). We also reproduce xresnet1d101 directly with 5 Kaiming initialization seeds, obtaining $0.947 \pm 0.013$ test macro AUC, lower and more variable than the published single-seed 0.957, with an outlier (seed 1337: 0.927, $1.5\sigma$ below mean) illustrating initialization instability.

DCT initialization produced $0.956 \pm 0.004$ macro AUC ($n\!=\!20$ batch seeds), matching the published xresnet1d101 benchmark (0.957) with more than 5$\times$ fewer parameters (Table~\ref{tab:main}). DCT initialization significantly exceeds Kaiming on the same architecture: $0.956 \pm 0.004$ vs $0.947 \pm 0.007$ ($p < 0.001$, Welch $t$-test, $n\!=\!20$ per group; Cohen's $d = 1.60$, 95\% CI $[0.89, 2.31]$). Kaiming variance is $2.8{\times}$ higher than DCT (std 0.007 vs 0.004), consistent with the Conformer pattern.

\subsection{Cross-Validation}\label{sec:cv}

\begin{figure}[!t]
\centering
\includegraphics[width=\linewidth]{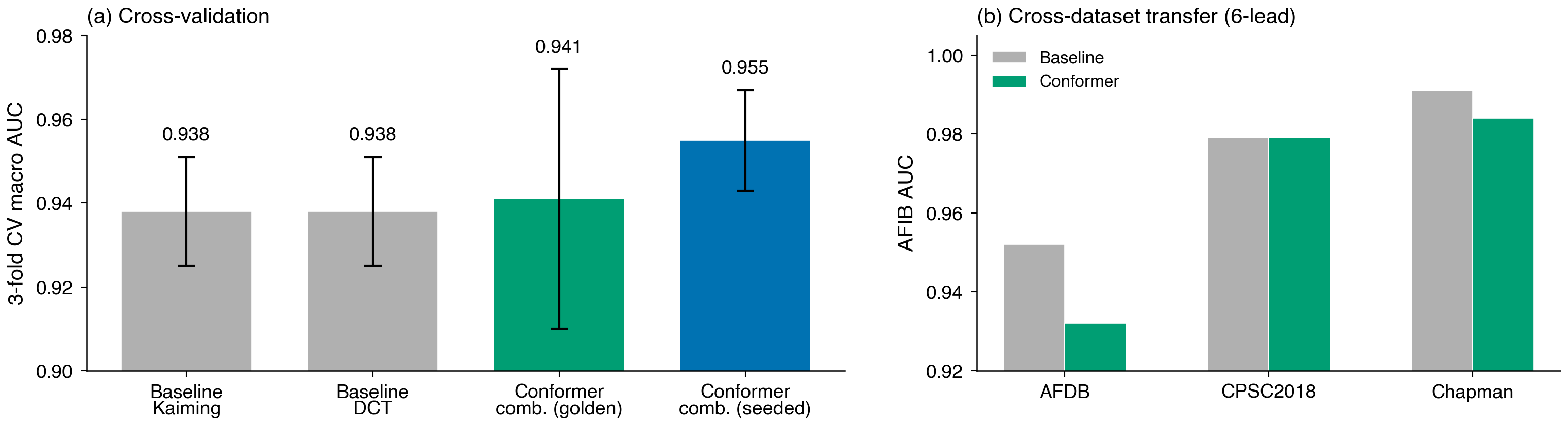}
\caption{Generalization validation. (a)~Three-fold cross-validation macro AUC for four configurations; Conformer combined ($0.955 \pm 0.012$) reaches the highest mean with seeded batch ordering. (b)~Cross-dataset AFIB detection AUC on three external databases (AFDB, CPSC2018, Chapman-Shaoxing): both architectures exceed 0.93 transfer AUC.}\label{fig:cv}
\end{figure}

Three-fold cross-validation (folds 10, 9, 8 as test sets) was performed on four configurations using strict determinism (Table~\ref{tab:cv}, Fig.~\ref{fig:cv}). All cross-validation standard deviations use the sample standard deviation ($n-1$ divisor). The Conformer combined with seeded batch ordering achieves $0.955 \pm 0.012$. The fully deterministic Conformer (golden ratio batching) drops to $0.941 \pm 0.031$, reflecting higher sensitivity to fold-specific data composition when batch ordering is data-dependent.

\begin{table}[!t]
\caption{Cross-validation results (3 folds, sample std). Each fold uses a single batch seed; Table~\ref{tab:main} multi-seed results average 20 batch orderings on fold~10. The golden fold~10 result (0.908) differs from Table~\ref{tab:main} (0.966) because CV golden uses a different validation fold for model selection, and golden ordering's data-dependence amplifies fold sensitivity (std=0.031 vs 0.012 for seeded).}\label{tab:cv}
\centering
\small
\begin{tabular}{lcccc}
\toprule
Configuration & F10 & F9 & F8 & Mean$\pm$Std \\
\midrule
Conform.\ mixed (seeded) & .942 & .961 & .963 & $.955\!\pm\!.012$ \\
Conform.\ mixed (golden) & .908 & .948 & .968 & $.941\!\pm\!.031$ \\
Baseline mixed & .927 & .953 & .934 & $.938\!\pm\!.013$ \\
Baseline Kaiming & .923 & .949 & .941 & $.938\!\pm\!.013$ \\
\bottomrule
\end{tabular}
\end{table}

Per-class CV analysis shows class-specific architecture preferences. The Conformer showed the highest CV mean for SARRH (0.941 vs 0.835 for the Kaiming baseline), the class with the highest Kaiming seed variance. The baseline showed the highest CV mean for PACE (0.973 vs 0.944 for Conformer). No single architecture dominates across all 12 classes.

TRIGU ($n=16$ total samples, 2 per test fold) exhibits high fold variance ($0.885 \pm 0.167$ for the fully deterministic Conformer), but this variance is validated across all three CV folds (each TRIGU sample tested exactly once) and confirmed by the 20-seed distribution on fold~10: TRIGU AUC range spans 30.9 percentage points under Kaiming versus 4.1 under structured initialization (Fig.~\ref{fig:perclass}). The high fold variance reflects genuine sensitivity to data composition, not measurement noise.

\subsection{Cross-Dataset Validation}\label{sec:crossdataset}

Two mixed-basis-initialized architectures were validated on three genuinely held-out external databases: the MIT-BIH Atrial Fibrillation Database (AFDB; 84,334 windows from long-term Holter recordings), the China Physiological Signal Challenge 2018 dataset (CPSC2018), and the Chapman-Shaoxing 12-lead ECG database (Table~\ref{tab:crossdataset}). All models were trained on PTB-XL using 6~common leads and evaluated zero-shot for AFIB detection. None of these datasets were used during development, architecture selection, or hyperparameter tuning.

\begin{table}[!t]
\caption{Cross-dataset generalization (6-lead models trained on 6 PTB-XL limb leads: I, II, III, aVR, aVL, aVF). Each entry is a single golden deterministic checkpoint. All models trained on PTB-XL with mixed-basis initialization and evaluated zero-shot for AFIB detection on three external databases. For AFDB, 4 of 6 leads are derived from 2 channels via Einthoven's law (Section~\ref{sec:data}). These 6-lead models are distinct from the 12-lead models used in all other experiments.}\label{tab:crossdataset}
\centering
\begin{tabular}{lllll}
\toprule
Model & AFDB & CPSC2018 & Chapman & Params \\
\midrule
Baseline (xresnet) & 0.952 & 0.979 & 0.991 & 1.65M \\
Conformer & 0.932 & 0.979 & 0.984 & 1.83M \\
\bottomrule
\end{tabular}
\end{table}

Both architectures achieve $>$0.93 AFIB AUC across all external databases, confirming generalization without data leakage. The baseline generalizes best on AFDB (0.952 vs 0.932 for Conformer), attributable to the wider 128-dimensional representation. On CPSC2018 and Chapman-Shaoxing, both models exceed 0.97~AFIB AUC. The Conformer's lower AFDB score (0.932) and more conservative prediction behavior (22.9\% AFIB positive rate vs 34.1\% for baseline) suggest that the self-attention mechanism may overfit to PTB-XL-specific R-R~interval distributions.

\section{Discussion}\label{sec:discussion}

Across two ECG architectures, structured orthogonal initialization significantly exceeds random Kaiming initialization ($n\!=\!20$ per group): Conformer ($p = 0.016$, $d = 0.82$) and Baseline CNN ($p < 0.001$, $d = 1.60$), with Kaiming exhibiting $2$--$3{\times}$ the variance in both cases. The $2{\times}2$ golden-seed design (Table~\ref{tab:golden}) isolates initialization from batch ordering, illustrating that Kaiming exhibits seed-dependent per-class variation on rare rhythms: trigeminy AUC drops to 0.781 (versus 0.980 with structured init). The basis comparison at $n{=}20$ (Table~\ref{tab:basis}) shows that basis choice does not significantly affect mean performance (Friedman $p{=}0.48$); the practical distinction is training variance, with DCT showing the highest (std=0.010) and sinusoidal the lowest (std=0.005). Any structured orthogonal basis produces Kaiming-equivalent performance with deterministic weights.

The critical distinction is between initialization determinism and full pipeline determinism. For initialization alone, the price is zero: structured initialization matches or exceeds Kaiming on every architecture and dataset tested, while providing seed-independent weights. For full pipeline determinism (including deterministic batch ordering), the price is real: the golden ratio Conformer yielded CV $0.941 \pm 0.031$ versus $0.955 \pm 0.012$ for seeded batching, a 1.4 percentage point loss with higher variance. We tested 11 seed-free batch ordering strategies (Appendix Table~\ref{tab:batch}); the best (class-guaranteed, 0.947) narrows the gap to 0.8 percentage points versus seeded shuffle (0.955), but requires class-label calibration.

The most consistent finding across all experiments is variance reduction, confirmed independently across domains. On ECG, structured initialization halves the aggregate variance (std 0.005 vs 0.010) and reduces per-class variability by up to $7.5{\times}$ (TRIGU range: 4.1pp vs 30.9pp, $n_\text{pos}\!=\!2$; SARRH: 2.5pp vs 4.2pp, $n_\text{pos}\!=\!77$; SVARR: 11.5pp vs 20.2pp, $n_\text{pos}\!=\!14$). On ChestMNIST (14 thorax diseases, multi-label, imbalanced), the same pattern emerges independently: DCT produces tighter per-class ranges in 11 of 14 classes, with the strongest effect on uncommon conditions (Pneumothorax: Kaiming range $3.7{\times}$ DCT; Emphysema: $2.4{\times}$; Fig.~\ref{fig:chestmnist}). On balanced benchmarks (BloodMNIST, OrganCMNIST, PathMNIST, CIFAR-100), structured initialization matches Kaiming with comparable variance. The pattern (variance reduction on imbalanced tasks, equivalence on balanced tasks) is confirmed independently in 1D (ECG) and 2D (chest X-ray), establishing that the benefit is a property of structured initialization itself, not task-specific.

Cross-domain validation on seven MedMNIST benchmarks ($n{=}20$ per condition, all $p > 0.14$; Table~\ref{tab:medmnist}) further confirms the ``no harm'' conclusion: structured initialization introduces no performance penalty on standard imaging tasks.

Jordan et al.~\cite{jordan2024calibrated} proved that test-set variance between training runs is mathematically inevitable for calibrated ensembles, arguing it is ``harmless'' because distribution-level metrics are stable. From the deep ensemble perspective, this inter-run variance enables uncertainty estimation and calibrated predictions. Our findings do not contradict this view. However, they qualify it for safety-critical single-model deployment: while aggregate metrics (macro AUC) vary by only 4.1 percentage points across 20 seeds, per-class metrics for rare classes swing by over 20 percentage points (Fig.~\ref{fig:perclass}). This variance is invisible in aggregate reporting but consequential for specific patient populations. When regulatory or deployment constraints require a single deterministic model rather than an ensemble, eliminating this variance becomes valuable.

The ablation (Table~\ref{tab:ablation}) measures batch-ordering variance ($n{=}5$ batch seeds per configuration) and should be interpreted cautiously. The DCT-only batch-ordering mean (0.948) falls within the DCT init-seed distribution (Table~\ref{tab:basis}: $0.956 \pm 0.010$, $n{=}20$), confirming that DCT initialization matches Kaiming in mean performance. The combined framework (mixed + cw, 0.961) numerically exceeds Kaiming (0.958) but does not reach statistical significance at $n\!=\!5$. At $n{=}20$, sqrt class weighting does not improve Kaiming: Kaiming+cw achieves $0.954 \pm 0.012$ ($n{=}18$ converged of 20; 2 seeds collapsed to $<$0.67) versus $0.953 \pm 0.010$ without cw ($p{=}0.82$, Welch $t$-test), while introducing a 10\% training failure rate absent from both Kaiming alone and all structured bases (0/80 failures). The $2{\times}2$ golden design (Table~\ref{tab:golden}), where both structured and Kaiming models use identical sqrt class weighting, shows a 3.1 percentage-point gap (0.959 vs 0.928), though this is a single deterministic observation, not a statistical comparison.

Structured initialization creates correlated gradients: all DCT filters are smooth cosines, so similar inputs produce similar activations across filters within a stage. This correlation contributes to DCT's higher training variance (std=0.010, matching Kaiming's 0.010, at $n{=}20$ init seeds). The $n{=}20$ basis comparison (Table~\ref{tab:basis}) shows a variance gradient across bases: DCT (0.010) $>$ Hadamard (0.008) $>$ Hartley (0.006) $\approx$ sinusoidal (0.005). While the pairwise variance differences do not reach significance (Levene's test, all $p > 0.72$), a largely concordant ordering appears in noise robustness rankings (Appendix Table~\ref{tab:noise}): Hartley $>$ Hadamard $>$ Sinusoidal $>$ DCT across all three noise types (sinusoidal and Hartley swap between variance and robustness rankings). The mixed-basis configuration reduces batch-ordering sensitivity by 4:1 (std 0.012 to 0.003, Table~\ref{tab:ablation}). Whether this reflects cross-stage decorrelation, properties of specific bases, or the dominance of the 23-block Hartley stage remains an open question.

Structured initialization maintains higher validation AUC with lower variance than Kaiming from epoch~20 onward (Fig.~\ref{fig:method}b), indicating that frequency-domain basis functions provide a more stable optimization trajectory on periodic physiological signals. The observation is consistent with recent findings by Fernandez-Hernandez et al.~\cite{miravet2025sinusoidal}, who report faster convergence with sinusoidal initialization across multiple architectures.

Beyond convergence speed, deterministic initialization guarantees spatial attribution stability: bit-identical weights produce bit-identical gradient-based explanations (e.g., GradCAM~\cite{selvaraju2017gradcam}), ensuring that spatial attributions are fixed and auditable across independent runs.

The practical implication is that deployment context should determine architecture choice. For maximum cross-dataset transfer, the unconstrained baseline with structured initialization provides both determinism and generalization. The Conformer provides self-attention maps for rhythm-level interpretability (which beat-to-beat intervals inform the prediction), though its cross-dataset performance (0.932 AFDB) trails the baseline (0.952).

\subsection{Limitations}\label{sec:limitations}

Structured initialization has been demonstrated on two convolutional ECG architectures and ResNet-18 on medical image benchmarks. Testing on non-convolutional architectures (RNNs, state space models) and pre-trained models would extend the generality claim. AFDB evaluation uses 2-channel recordings with 6 limb leads derived via Einthoven's law, a distribution shift from PTB-XL training since precordial lead information is absent. The mixed-basis stage assignment was selected heuristically; a systematic search over basis permutations has not been performed. All experiments use batch size 128; the interaction between structured initialization's correlated gradients and batch size has not been explored.

Structured initialization eliminates initialization randomness but not batch ordering randomness. Golden ratio ordering achieves full determinism at a performance cost: among 11 seed-free strategies evaluated (Appendix Table~\ref{tab:batch}), the best (class-guaranteed, 0.947) narrows the gap to 0.8 percentage points versus seeded shuffle (0.955) but requires class-label calibration. Closing this gap with a parameter-free method remains open.

Clinical validation requires prospective studies with cardiologist-adjudicated outcomes, following reporting standards such as TRIPOD+AI~\cite{collins2024tripod}. Demographic subgroup analysis ($n{=}20$ seeds, 9 rhythm classes evaluable in both sex groups, excluding SVTAC, PSVT, and TRIGU with $\leq 2$ test positives per sex) shows no significant sex disparity (permutation $p = 0.85$). Independent reproduction by other groups is essential before any consideration for potential clinical applications.

\section{Conclusion}\label{sec:conclusion}

We have presented a framework for verified bit-identical deep learning training that systematically eliminates three sources of non-determinism: weight initialization (via structured orthogonal basis functions), batch ordering (via golden ratio batch ordering), and non-deterministic GPU operations (via architecture selection). The framework produces trained models with MD5-verified identical weights across independent runs.

On PTB-XL ECG rhythm classification, structured initialization significantly exceeds Kaiming on both architectures tested (Table~\ref{tab:main}): the Conformer ($p = 0.016$, $d = 0.82$, $n\!=\!20$) with Kaiming exhibiting twice the variance, and the baseline CNN ($p < 0.001$, $d = 1.60$, $n\!=\!20$) with Kaiming exhibiting $2.8{\times}$ the variance. A four-basis comparison at $n{=}20$ (Table~\ref{tab:basis}) showed that all structured orthogonal bases produce equivalent mean performance (Friedman $p{=}0.48$), with the practical distinction among bases being training variance (DCT std=0.010, sinusoidal std=0.005), not mean performance. Variance decomposition (Conformer) estimates initialization variance (zero by construction), batch-ordering variance (std\,=\,0.005), and fold variance (std\,=\,0.012), enabling quantification of each source.

Cross-domain validation on seven MedMNIST benchmarks ($n\!=\!20$ per condition, all $p > 0.14$; Table~\ref{tab:medmnist}) confirms no performance penalty on standard tasks, while per-class analysis on ChestMNIST independently replicates the ECG variance reduction pattern on rare classes (11 of 14 classes, Pneumothorax $3.7{\times}$). Structured initialization matters most on imbalanced tasks with rare classes, precisely where seed-dependent failures are clinically consequential. Cross-dataset evaluation on three external ECG databases not used during model development confirms generalization, with both architectures exceeding 0.93 AFIB AUC (Table~\ref{tab:crossdataset}).

The framework provides a methodological tool for algorithmic reproducibility in deep learning for medical applications. As regulatory frameworks increasingly value transparency and process documentation for AI/ML-based medical devices~\cite{fda2021aiml,euaiact2024}, methods that eliminate non-determinism at the algorithmic level, instead of merely fixing random seeds, facilitate verifiable training pipelines. All code, evaluation scripts, and experimental evidence are publicly available. Trained checkpoints are reproducible from the released code by construction.

\aimonly{
\section*{CRediT authorship contribution statement}

\textbf{Yakov P. Shkolnikov}: Conceptualization, Methodology, Software, Validation, Formal analysis, Investigation, Data curation, Writing -- original draft, Writing -- review \& editing, Visualization.
}

\section*{Declaration of competing interests}

The author declares no competing interests.

\section*{Funding}

This research did not receive any specific grant from funding agencies in the public, commercial, or not-for-profit sectors.

\section*{Ethics statement}

This study uses only publicly available, de-identified datasets (PTB-XL, MIT-BIH AFDB, CPSC2018, Chapman-Shaoxing, MedMNIST, CIFAR) and does not involve human subjects research. No ethics approval was required.

\section*{Declaration of generative AI and AI-assisted technologies in the manuscript preparation process}

During the preparation of this work the author used Claude (Anthropic) in order to assist with code development, manuscript drafting, and editorial revision. After using this tool, the author reviewed and edited the content as needed and takes full responsibility for the content of the published article.

\section*{Data and code availability}

PTB-XL v1.0.3 is available at PhysioNet~\cite{goldberger2000physionet} (\url{https://doi.org/10.13026/6sec-a640}). The MIT-BIH Atrial Fibrillation Database is available at PhysioNet (\url{https://doi.org/10.13026/C2FS0H}). The MIT-BIH Noise Stress Test Database is available at PhysioNet (\url{https://doi.org/10.13026/C2HS-1990}). CPSC2018~\cite{cpsc2018} is available at \url{http://2018.icbeb.org/Challenge.html}. Chapman-Shaoxing~\cite{zheng2020chapman} is available at \url{https://doi.org/10.6084/m9.figshare.c.4560497}. MedMNIST v2~\cite{yang2023medmnist} is available at \url{https://medmnist.com/}. All datasets used in this study are publicly available and require no access restrictions. All code, evaluation scripts, and experimental evidence are available at \url{https://github.com/yshk-mxim/repnet}.


\bibliographystyle{elsarticle-num}
\bibliography{references}


\appendix

\section{CIFAR Image Classification Results}\label{app:cifar}

To validate DCT initialization beyond medical domains, we applied 2D-DCT initialization to ResNet-18 on CIFAR-100 image classification~\cite{he2016deep}. 2D-DCT initialization uses separable 2D-DCT basis functions computed over the ($k_h$, $k_w$) kernel dimensions, with 1D-DCT for channel mixing. Data augmentation uses per-epoch seeding (\texttt{torch.manual\_seed(epoch $\times$ 1000)}), making all augmentation operations deterministic. Sample indices replace the signal hash for golden ratio ordering: $\text{key}[i] = \text{mod}(i \cdot \varphi + \text{epoch} \cdot \varphi, 1)$.

\begin{table}[!t]
\caption{CIFAR-100 image classification results (ResNet-18, 20 seeds). DCT = single-basis 2D-DCT initialization. TOST equivalence confirmed at $\delta\!=\!0.5$ percentage points ($p < 0.001$).}\label{tab:cifar}
\centering
\small
\begin{tabular}{llcr}
\toprule
Dataset & Init & Accuracy & Params \\
\midrule
CIFAR-100 & DCT 2D & $78.4 \pm 0.3$\% & 11.2M \\
CIFAR-100 & Kaiming & $78.3 \pm 0.4$\% & 11.2M \\
\bottomrule
\end{tabular}
\end{table}

With 20 seeds per condition (Table~\ref{tab:cifar}), DCT and Kaiming are statistically equivalent on CIFAR-100 ($78.4 \pm 0.3$\% vs $78.3 \pm 0.4$\%; $t = 0.77$, $p = 0.45$; Cohen's $d = 0.24$; 95\% CI $[-0.14, +0.31]$ pp). TOST equivalence testing confirms equivalence at $\delta = 0.5$ percentage points ($p < 0.001$), providing definitive evidence that DCT initialization matches Kaiming on a standard benchmark with proper statistical power. DCT also exhibits lower seed variance ($\sigma = 0.26$ vs $0.42$). MedMNIST medical image results (Table~\ref{tab:medmnist}) further validate the approach on clinical imaging tasks.

\section{Noise Robustness Details}\label{app:noise}

\begin{table}[!t]
\caption{Noise robustness by orthogonal basis (Conformer, 12-lead PTB-XL, $n{=}5$ seeds per basis, 0\,dB SNR). Each column shows AUC drop from clean at the harshest noise level. Friedman tests: baseline wander $p{=}0.18$, muscle artifact $p{=}0.18$, electrode motion $p{=}0.069$; none reach significance at $\alpha{=}0.05$, though the electrode motion ordering approaches significance. Noise sources from the MIT-BIH Noise Stress Test Database~\cite{moody1984nstdb}.}\label{tab:noise}
\centering
\footnotesize
\begin{tabular}{lcccc}
\toprule
Basis & Clean AUC & BW drop & Muscle drop & EM drop \\
\midrule
Hartley     & $\mathbf{.959 \pm .003}$ & $\mathbf{.033 \pm .012}$ & $\mathbf{.035 \pm .010}$ & $\mathbf{.057 \pm .019}$ \\
Hadamard    & $.957 \pm .008$ & $.041 \pm .019$ & $.046 \pm .023$ & $.070 \pm .022$ \\
Sinusoidal  & $.951 \pm .006$ & $.050 \pm .011$ & $.063 \pm .021$ & $.064 \pm .019$ \\
DCT         & $.951 \pm .005$ & $.054 \pm .009$ & $.063 \pm .017$ & $.091 \pm .017$ \\
\bottomrule
\end{tabular}
\end{table}

At $n{=}5$, Hartley shows the lowest noise degradation across all three noise types (Table~\ref{tab:noise}). DCT shows the largest electrode motion drop ($0.091 \pm 0.017$), while Hartley loses $0.057 \pm 0.019$. Friedman tests show no significant basis effect for baseline wander ($p{=}0.18$) or muscle artifact ($p{=}0.18$), but electrode motion approaches significance ($\chi^2{=}7.08$, $p{=}0.069$), where the Hartley $>$ Hadamard $>$ Sinusoidal $>$ DCT ordering is most pronounced. The pattern is broadly consistent with the clean-AUC ranking (Table~\ref{tab:basis}). Baseline wander (low-frequency drift), muscle artifact (high-frequency EMG contamination), and electrode motion (abrupt signal discontinuities) all show the same directional ordering.

\section{Batch Ordering Strategies}\label{app:batch}

\begin{table}[!t]
\caption{Seed-free batch ordering strategies (Conformer, mixed-basis init, PTB-XL fold~10, single run each). All seed-free methods produce deterministic batch sequences without any random seed. Golden ratio was selected for simplicity (no hyperparameters, no class-label dependence) despite lower AUC than class-guaranteed ordering.}\label{tab:batch}
\centering
\footnotesize
\begin{tabular}{@{}lccp{4.5cm}@{}}
\toprule
Method & AUC & Seed-free? & Notes \\
\midrule
Seeded shuffle (control) & 0.955 & No & Upper bound \\
\midrule
Class guaranteed & \textbf{0.947} & Yes & Highest seed-free; needs calibration \\
Golden ratio & 0.941 & Yes & Selected: parameter-free \\
Sobol sequence & 0.939 & Yes & Quasi-random coverage \\
Feature-space diverse & 0.883 & Yes & Rotating projection \\
Stratified quantile & 0.579 & Yes & Loss-based, class-correlated \\
Class diverse & 0.559 & Yes & Class interleave \\
Herding & 0.522 & Yes & Static order, overfits \\
Content hash (SHA-256) & 0.479 & Yes & Hash correlates with class \\
PairBalance (GraB) & 0.579 & Yes & Loss-based ordering \\
Loss-ranked & 0.500 & Yes & Homogeneous batches \\
Interleaved loss-ranked & 0.579 & Yes & Still class-correlated \\
\bottomrule
\end{tabular}
\end{table}

Class-guaranteed ordering (0.947) outperforms golden ratio (0.941) by AUC but requires specifying a minimum per-batch class count, making it label-dependent and dataset-specific. Golden ratio ordering was selected for the main experiments because it is entirely parameter-free: the permutation depends only on signal content and epoch index, requires no class labels or calibration, and transfers directly to new datasets. Sobol-sequence ordering performs comparably (0.939).

Loss-based methods (stratified quantile, PairBalance, loss-ranked, interleaved loss-ranked) all fail catastrophically because in multi-label ECG classification, loss correlates strongly with class membership: the dominant SR class (77\% of samples) is consistently low-loss, while rare arrhythmias are high-loss. Any loss-based ordering implicitly sorts by class, producing homogeneous batches. The golden ratio method avoids this by computing permutations from signal content rather than training dynamics.

\section{Theoretical Analysis of DCT Initialization}\label{app:theory}

This section analyzes the information-theoretic properties of DCT-initialized convolutional layers compared to random (Kaiming) initialization. The notation $C_{k,n}$ used below corresponds to $d_i[j]$ in the implementation description (Eq.~\eqref{eq:dct}).
Consider a convolutional layer $\mathbf{y} = \mathbf{W}\mathbf{x}$, where $\mathbf{x} \in \mathbb{R}^N$ is the input signal and $\mathbf{W} \in \mathbb{R}^{M \times N}$ is the weight matrix.

\textbf{Definition (DCT-II Weight Matrix).}
The orthonormal DCT-II matrix $\mathbf{C} \in \mathbb{R}^{N \times N}$ has entries
\begin{equation}
C_{k,n} = \alpha_k \cos\!\left(\frac{\pi(n + \tfrac{1}{2})k}{N}\right), \quad \alpha_k = \begin{cases} \sqrt{1/N} & k = 0 \\ \sqrt{2/N} & k \geq 1 \end{cases}
\end{equation}
satisfying $\mathbf{C}^{\!\top}\mathbf{C} = \mathbf{C}\mathbf{C}^{\!\top} = \mathbf{I}_N$. For a Conv1d($C_\text{in}$, $C_\text{out}$, $K$), we set $\mathbf{W}_\text{DCT}$ to the first $C_\text{out}$ rows of the DCT-II matrix in the $(C_\text{in} \cdot K)$-dimensional space, scaled to match Kaiming variance.

\textbf{Proposition 1 (Energy Preservation Bound).}
Let $\mathbf{x} \in \mathbb{R}^N$ be an arbitrary signal and let $\mathbf{W}_K$ consist of $K \leq N$ rows of the DCT-II matrix. Then
\begin{equation}
\frac{\|\mathbf{W}_K \mathbf{x}\|^2}{\|\mathbf{x}\|^2} = \frac{\sum_{k=0}^{K-1} |\langle \mathbf{c}_k, \mathbf{x} \rangle|^2}{\|\mathbf{x}\|^2} \geq 1 - \frac{\sum_{k=K}^{N-1} |\hat{x}_k|^2}{\|\mathbf{x}\|^2}
\end{equation}
where $\hat{x}_k = \langle \mathbf{c}_k, \mathbf{x}\rangle$ are the DCT coefficients. For the full matrix ($K = N$), Parseval's theorem gives $\|\mathbf{C}\mathbf{x}\|^2 = \|\mathbf{x}\|^2$ exactly.
For Kaiming initialization $\mathbf{W} \sim \mathcal{N}(\mathbf{0}, \frac{2}{N}\mathbf{I})$, the expected energy ratio is $\mathbb{E}[\|\mathbf{W}\mathbf{x}\|^2/\|\mathbf{x}\|^2] = 2K/N$ with variance $8K/N^2$.

\textit{Proof.} The bound follows from Parseval's identity: $\|\mathbf{x}\|^2 = \sum_{k=0}^{N-1}|\hat{x}_k|^2$. For Kaiming, each $W_{ij} \sim \mathcal{N}(0, 2/N)$, so $(\mathbf{W}\mathbf{x})_k \sim \mathcal{N}(0, 2\|\mathbf{x}\|^2/N)$, giving a $\chi^2_K$-distributed squared norm with the stated moments. \hfill$\square$

\textbf{Theorem 1 (Condition Number Bound).}
The DCT-II matrix satisfies $\kappa(\mathbf{C}) = 1$ (all singular values equal~1). For Kaiming $\mathbf{W} \in \mathbb{R}^{N \times N}$ with i.i.d.\ $\mathcal{N}(0, \sigma^2)$ entries, the Marchenko--Pastur law~\cite{marchenko1967distribution} gives:
\begin{equation}
\kappa(\mathbf{W}) \to \frac{1+\sqrt{\gamma}}{1-\sqrt{\gamma}} \quad \text{as } N \to \infty, \quad \gamma = \frac{M}{N}.
\end{equation}
For square matrices ($\gamma = 1$), $\kappa \to \infty$. Since $\nabla_\mathbf{x}\mathcal{L} = \mathbf{W}^\top \nabla_\mathbf{y}\mathcal{L}$, gradient component ratios are bounded by $\kappa(\mathbf{W})$.

\textbf{Corollary (Band-Limited Signals).}
For ECG signals sampled at $f_s = 100$\,Hz with diagnostic content below 40\,Hz ($B = 0.8$), the first $\lceil 0.8N \rceil$ DCT coefficients capture $> 99\%$ of signal energy. In a Conv1d layer with kernel size $K = 5$, the full $5 \times 5$ DCT matrix captures all within-window energy with $\kappa = 1$.

\subsection{Connection to Optimal Decorrelation}

\begin{figure}[!t]
\centering
\includegraphics[width=\linewidth]{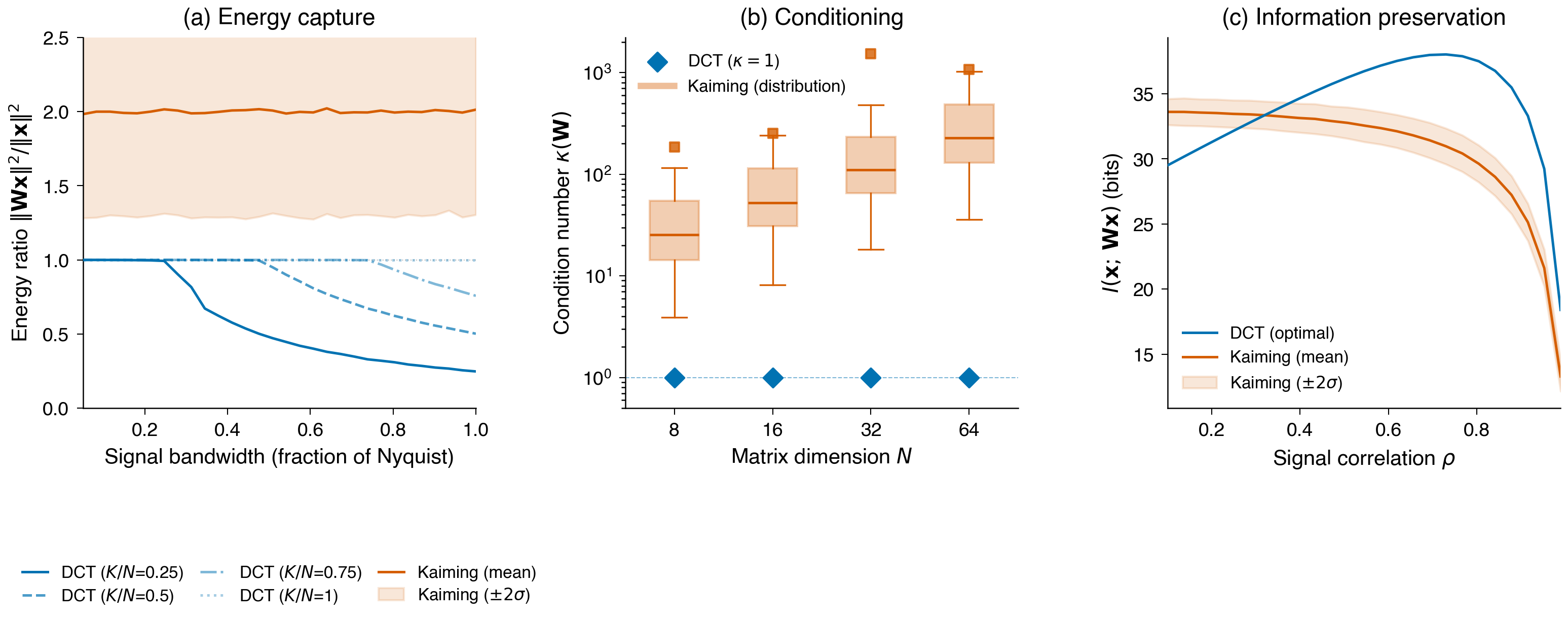}
\caption{Theoretical analysis of DCT vs.\ Kaiming initialization. (a)~Energy capture ratio for band-limited signals ($N\!=\!64$): DCT with full rank preserves energy exactly; Kaiming has correct expectation but high variance. (b)~Condition number: DCT is always 1 (optimal gradient flow); Kaiming grows with matrix dimension $N$. (c)~Mutual information $I(\mathbf{x};\, \mathbf{Wx})$ for a dimensionality-reducing layer ($M\!=\!N/4$, $N\!=\!64$, SNR\,=\,10\,dB) with Toeplitz-correlated input: for correlated signals ($\rho \gtrsim 0.3$), DCT preserves more mutual information than Kaiming by selecting the most informative spectral components; at low correlation, Kaiming's higher per-filter variance (fan-in scaling) compensates for its random projection directions.}\label{fig:dct_theory}
\end{figure}

The DCT-II is the asymptotically optimal transform for decorrelating first-order Markov processes~\cite{rao1990discrete}. ECG signals are well-modeled as quasi-stationary with autoregressive structure, making the DCT a near-optimal linear transform in the Karhunen--Lo\`eve sense:
\begin{equation}
\mathbf{C}\boldsymbol{\Sigma}_x\mathbf{C}^\top \approx \text{diag}(\lambda_0, \ldots, \lambda_{N-1}).
\end{equation}
This near-diagonalization means DCT initialization approximately whitens the input, ensuring each output channel captures an independent spectral component (Fig.~\ref{fig:dct_theory}).

\end{document}